\documentclass[acmsmall]{acmart}
\AtBeginDocument{%
  }

\setcopyright{cc}
\acmJournal{CSUR}
\acmDOI{10.1145/3834764}

\usepackage[utf8]{inputenc}
\usepackage{textcomp}
\usepackage{stfloats}
\usepackage{url}
\usepackage{verbatim}
\usepackage{subfigure}
\usepackage{graphicx}
\usepackage{booktabs}
\usepackage{multirow}
\usepackage{hyperref}
\usepackage{tabularx}
\usepackage{pifont}
\usepackage[table]{xcolor}
\usepackage{tikz_commands}

 \usepackage{cleveref} 
 \crefname{section}{Section}{Sections}
 \crefname{theorem}{Theorem}{Theorems}
 \crefname{lemma}{Lemma}{Lemmas}
 \crefname{equation}{Equation}{Equations}
 \crefname{proposition}{Proposition}{Propositions}
 \crefname{claim}{Claim}{Claims}
\crefname{appendix}{Appendix}{Appendices}
   \crefname{algorithm}{Algorithm}{Algorithms}
 \crefname{figure}{Figure}{Figures}
 \crefname{table}{Table}{Tables}
 \crefname{remark}{Remark}{Remarks}
 \crefname{definition}{Definition}{Definitions}
 \crefname{equatinon}{Equation}{Equations}
 \crefname{corollary}{Corollary}{Corollaries}

\allowdisplaybreaks

\usepackage{thmtools}
\usepackage{thm-restate}

\let \oldtextcircled \textcircled
\renewcommand{\textcircled}[1]{\oldtextcircled{\footnotesize #1}}

\usepackage{enumitem}
\setlist[itemize]{leftmargin=9mm}

\def \v{\mathbf{v}}

\def \b{\mathbf{b}}
\def \d{\mathbf{d}}
\def \x{\mathbf{x}}
\def \y{\mathbf{y}}

\def \e{\mathbf{e}}

\def \z{\mathbf{z}}

\def \v{\mathbf{v}}

\def \BK{\mathbf{K}}
\def \BA{\mathbf{A}}

\def \BX{\mathbf{X}}
\def \BY{\mathbf{Y}}
\def \BI{\mathbf{I}}

\def \BD{\mathbf{D}}
\def \BE{\mathbf{E}}

\def \BH{\mathbf{H}}
\def \BL{\mathbf{L}}

\def \BM{\mathbf{M}}
\def \BG{\mathbf{G}}
\def \BP{\mathbf{P}}
\def \BQ{\mathbf{Q}}
\def \BR{\mathbf{R}}

\def \BT{\mathbf{T}}
\def \BU{\mathbf{U}}
\def \BV{\mathbf{V}}
\def \BW{\mathbf{W}}
\def \BZ{\mathbf{Z}}

\def \btheta{\bm{\theta}}

\usepackage{mathtools}

\def \m{\mathbf{m}}

\def \btheta{\boldsymbol{\theta}}

\newcommand{\cmark}{\ding{51}}

\begin{document}

\title{A Survey of Graph Transformers: Architectures, Theories and Applications}

\author{Chaohao Yuan}
\affiliation{%
  \institution{The Chinese University of Hong Kong}
  \city{Hong Kong S.A.R.}
  \country{China}}
\affiliation{%
  \institution{SIGS, Tsinghua University}
  \city{Shenzhen}
  \country{China}}
\email{chaohaoyuan@link.cuhk.edu.hk}

\author{Kangfei Zhao}
\affiliation{%
  \institution{The Chinese University of Hong Kong}
  \city{Hong Kong S.A.R.}
  \country{China}
}
\email{zkf1105@gmail.com}

\author{Ercan Engin Kuruoglu}
\affiliation{%
 \institution{SIGS, Tsinghua University}
 \city{Shenzhen}
 \country{China}
}
\affiliation{%
 \institution{ISTI-CNR}
 \city{Pisa}
 \country{Italy}
}
\email{kuruoglu@sz.tsinghua.edu.cn}

\author{Liang Wang}
\affiliation{%
  \institution{Institute of Automation, Chinese Academy of Sciences}
  \city{Beijing}
  \country{China}}
\email{liang.wang@cripac.ia.ac.cn}

\author{Tingyang Xu}
\affiliation{%
  \institution{DAMO Academy, Alibaba Group}
  \city{Hangzhou}
  \country{China}}
\affiliation{%
  \institution{Hupan Lab}
  \city{Hangzhou}
  \country{China}}
\email{xuty\_007@hotmail.com}

\author{Wenbing Huang}
\affiliation{%
  \institution{Renmin University of China}
  \city{Beijing}
  \country{China}}
\email{hwenbing@126.com}

\author{Deli Zhao}
\affiliation{%
  \institution{DAMO Academy, Alibaba Group}
  \city{Hangzhou}
  \country{China}}
\affiliation{%
  \institution{Hupan Lab}
  \city{Hangzhou}
  \country{China}}
\email{zhaodeli@gmail.com}

\author{Hong Cheng}
\affiliation{%
  \institution{The Chinese University of Hong Kong}
  \city{Hong Kong S.A.R.}
  \country{China}}
\email{hcheng@se.cuhk.edu.hk}

\author{Yu Rong}
\authornote{Yu Rong is the corresponding author.}
\affiliation{%
  \institution{DAMO Academy, Alibaba Group}
  \city{Hangzhou}
  \country{China}}
\affiliation{%
  \institution{Hupan Lab}
  \city{Hangzhou}
  \country{China}}
\email{yu.rong@hotmail.com}

\renewcommand{\shortauthors}{C. Yuan et al.}

\begin{abstract}
Graph Transformers (GTs) have demonstrated a strong capability in modeling graph structures by addressing the intrinsic limitations of graph neural networks (GNNs), such as over-smoothing and over-squashing. Recent studies have proposed diverse architectures, enhanced explainability, and practical applications for Graph Transformers. In light of these rapid developments, we conduct a comprehensive review of Graph Transformers, covering aspects such as their architectures, theoretical foundations, and applications. In this survey, we first categorize the architecture of Graph Transformers according to their strategies for processing structural information, including graph tokenization, positional encoding, structure-aware attention, and model ensemble. Then, from the theoretical perspective, we examine the expressivity of Graph Transformers in various discussed architectures and contrast them with other advanced graph learning algorithms to discover their connections. For applications, we organize the literature around four graph organization forms, from relational, geometric, dynamic to heterogeneous. A Practical Guidance table then maps architectural components to these graph forms by adoption frequency, so practitioners can narrow down which design families to consider for a given input structure. Lastly, we will discuss the current challenges and prospective directions in Graph Transformers for potential future research.
\end{abstract}




\maketitle

\section{Introduction}

Graph data, a non-Euclidean data structure, is commonly found in various real-world applications, including molecular data, protein interactions, and social networks. Recently, Graph Neural Networks (GNNs)~\cite{kipf2016semi} have demonstrated impressive capabilities in modeling such data. A representative paradigm for building GNNs is the message-passing~\cite{gilmer2017neural}, which iteratively aggregates the neighbours' information and updates the node embedding.
Nonetheless, the message-passing paradigm encounters several intrinsic limitations, such as over-smoothing~\cite{rong2019dropedge,chen2020measuring} and over-squashing~\cite{kreuzer2021rethinking}, making it challenging for GNNs to effectively capture the long-range dependencies within the graphs. 

In another line of research, the Transformer model~\cite{vaswani2017attention} has demonstrated remarkable performance across a range of modalities, including natural language~\cite{vaswani2017attention}, images~\cite{dosovitskiy2021an}, videos~\cite{dosovitskiy2021an}, and time-series~\cite{time-series-survey}. A notable advantage of the Transformer model is its ability to effectively capture long-range dependencies, making it a feasible solution for addressing the limitations inherent in GNNs. Graph Transformers (GTs) adapt Transformer architecture to handle both node embeddings and graph structures, demonstrating superior performance compared to message-passing GNNs on a variety of tasks, including the prediction of molecular properties~\cite{rong2020self, Graphormer, tholke2022torchmd}, molecular dynamics simulation~\cite{fuchs2020se3transformers,liao2023equiformer, equiformer_v2}, and graph generation~\cite{vignac2023digress}.

In this survey, we conduct a systematic and comprehensive review of the recent advancements in GTs, examining the developments from the perspectives of architectures, theories, and applications. 
First, from the perspective of architecture, we categorize GTs into four categories based on their ways of integrating graph structures into Transformers. 
{(1) Multi-level Graph Tokenization.} These models utilize tokenization mechanisms to represent edges, subgraphs, and hops as structure-aware tokens, enabling the attention mechanism to effectively capture and learn intrinsic topological relationships.
{(2) Structural Positional Encoding.} These models enhance positional encoding (PE), traditionally used to denote token sequence relationships, to elucidate the structural interrelations among tokens.
{(3) Structure-aware Attention Mechanisms.} 
Since the attention matrix inherently captures the learned relationships between tokens, these models modify the attention matrix using graph-derived structural information to incorporate node interrelations.
{(4) Model Ensemble between GNNs and Transformers}. In addition to the direct modification of the architectures of Transformers, another effective strategy employs message-passing GNNs to encode structural information, followed by the integration of GNNs with Transformers. We follow the taxonomy of the previous study~\cite{min2022transformer} to organize this category. 
Additionally,  we investigate two critical aspects of Graph Transformer architecture: scalability challenges in large-scale graph processing and specialized architectural designs for geometric graphs.

After reviewing the architectures of GTs, it is important to determine which architecture is more powerful in general or in a specific task, since different architectures include different inductive biases or model the graph in different granularity. 
To this end, we investigate the expressive ability of GTs. 
Specifically, we examine the studies that compare the expressivity of GTs using the Weisfeiler-Lehman Test. These theoretical insights will enhance our understanding of how each component contributes to GTs' ability to learn graph data. Additionally, we discuss the relationships between GTs and other current graph learning algorithms, which can further elucidate the strengths and weaknesses of GTs.

\begin{figure}
    \centering
    \includegraphics[width=0.95\linewidth]{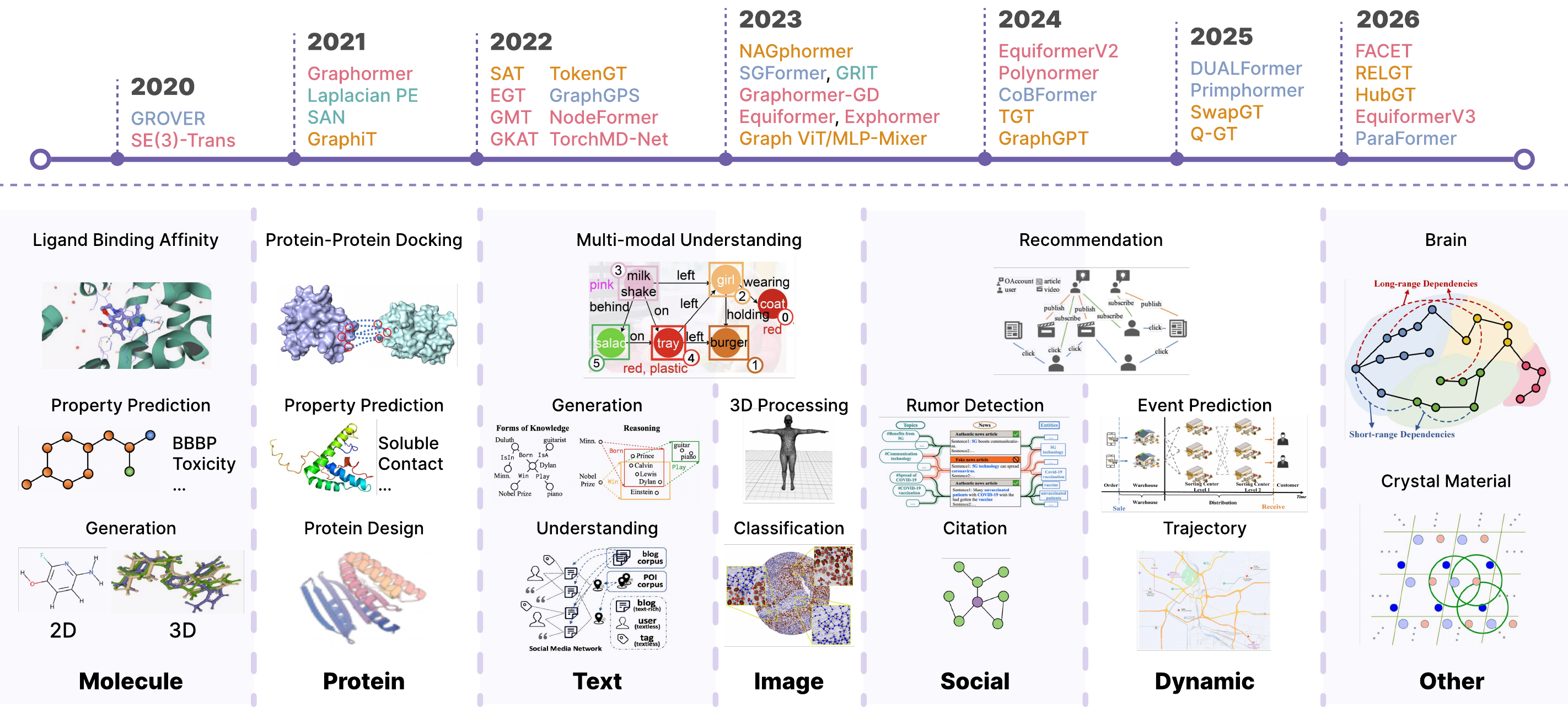}
    \vspace{-3ex}
    \caption{The above figure shows the development of the Graph Transformer architectures, where \textcolor[HTML]{8799BF}{blue} are the ensemble of GNNs and Transformers, and \textcolor[HTML]{CD7583}{red} are structure-aware attention, \textcolor[HTML]{7BAEAE}{green} are structural positional encoding and \textcolor[HTML]{FBB82A}{yellow} are multi-level tokenization. The below figure illustrates representative GT applications under four graph organization forms, including relational, geometric, dynamic, and multimodal graphs. The illustrations of the application examples are adapted from previous studies~\cite{rong2020self, zhou2023unimol, wu2024ebmdock, vignac2023digress, zheng2023structure, jin2023heterformer, kgtransformer, nakhli2023sparse, lin2021mesh, he2023multimodal, chen2024leveraging, zhang2024heterogeneous, wang2025hpst, jin2025llgformer, min2022masked, yu2024longrange, yan2022periodic}. 
}
    \label{fig:applications}
    \vspace{-5ex}
\end{figure}

Furthermore, to review GTs tailored for specific tasks, we organize their applications into four graph organization forms: relational graphs with discrete semantics, geometric and periodic graphs, dynamic graphs and event streams, and heterogeneous or multimodal graphs. To be more specific, the concrete representative examples are illustrated in \cref{fig:applications}.
For each organization form, we review the common task templates, dominant structural signals, and GT design choices that repeatedly appear across different domains. Such an organization avoids repeatedly introducing similar task definitions for each domain and helps connect applications with the corresponding GT architectures. In addition, applying GTs to specific tasks may require extra pre-training strategies or optimization objectives, such as diffusion. These aspects are also included in our review. In the last part of this survey, we discuss potential future directions of GT.

This survey aims to provide a thorough analysis of GTs from multiple perspectives, including their model architectures, expressivity, and applications.  While recent reviews have mainly focused on architectural aspects~\cite{min2022transformer, muller2024attending}, our work distinguishes itself within three aspects: (1) comprehensive architectural insights, (2) systematic analysis of theoretical expressivity of GTs, (3) extensive investigation of cross-domain applications. More importantly, these three perspectives are connected through the unified framework in \cref{fig:framework}, which provides a more systematic way to understand GT design choices and their corresponding capabilities. We also explore prospective research directions with recent advances in GTs.

\textbf{Roadmap.} The rest of this paper is organized as follows. Section~\ref{sec.prelimiary} introduces the preliminaries. 
Section~\ref{sec.architectures} reviews the four categories of GTs from the perspective of architecture. 
In Section~\ref{sec.theory}, we summarize the expressive capability of GTs and connect them to graph learning methodologies. 
Section~\ref{sec.applications} reviews the applications of GTs. 
We discuss the current situation of graph learning and point out the future research directions in Section~\ref{sec.future} and conclude the paper in Section~\ref{sec.conclusion}. 

\section{Preliminaries}
\label{sec.prelimiary}

In this section, we provide a concise overview of the essential notations pertaining to GTs, as well as a summary of the fundamental architecture of the vanilla Transformer and message passing neural networks.

We denote a graph as $\mathcal G = (\mathcal V, \mathcal E)$, where $\mathcal V$ represents the set of nodes and $\mathcal E$ represents the set of edges. The node set $\mathcal V$ comprises $n$ nodes, with the feature matrix $\BH \in \mathbb{R}^{n\times d}$, where $n$ is the number of nodes and $d$ is the dimension of the node features. The edge set $\mathcal E$ corresponds to an adjacency matrix $\mathbf{A}^{\BG} \in \mathbb R^{n \times n}$, where $\mathbf{A}^{\BG}_{u,v} = 1$ if there exists an edge $(u, v)$ in $\mathcal E$, and $\mathbf{A}^{\BG}_{u,v} = 0$ otherwise. 

The aforementioned notations are adequate for representing a basic graph with an adjacency matrix and node features. In more challenging scenarios, graphs are associated with more features to support various applications 1): Geometric Graph: In addition to node property features, the nodes will have coordinate features $\Vec{\BX} \in \mathbb{R}^{3}$, which will be independently handled to achieve invariance or equivariance properties.
2): Graph with edge features: Rather than merely converting the edge information into an adjacency matrix, the edges offer supplementary information that denotes more specific relationships between the nodes.
3): Dynamic graph: The graph additionally contains the time domain $\mathcal T$ compared with static graphs, represented as $\mathcal G = (\mathcal V, \mathcal E, \mathcal T)$. This indicates that the nodes and edges can change over time. The graph is denoted as $\mathcal G = \{(v_i, v_j, \tau)_n, n = 1, 2, \ldots, |\mathcal E|\}$, where each tuple $(v_i, v_j, \tau)$ represents an edge between node $v_i$ and node $v_j$ at a specific time $\tau \in \mathcal T$ and $\mathcal E$ specifies the edge set for that particular timestep.


\textbf{Message Passing Graph Neural Networks}: 
Graph neural networks~\cite{kipf2016semi} are a foundational framework for learning representations in graphs via the message passing mechanism~\cite{gilmer2017neural}. Given node embeddings $\BH$ and the adjacency matrix $\BA^{\BG}$, a message passing neural network (MPNN) $\phi_\theta$ updates  node embeddings via the propagating information from neighboring nodes.
The process of modeling node embeddings in an MPNN can be formally expressed as:
\begin{equation}
    \vspace{-1ex}
    \m_{ij} = \phi_{msg}(\BH_i, \BH_j, \e_{ij}),
    \quad
    \BH_{i} = \phi_{upd}(\BH_i, \{\m_{ij} \}_{j \in \mathcal{N}_i}),
\end{equation}

where $\BH$ and $\e$ represent node and edge embeddings, respectively. Here, $\phi_{msg}(\cdot)$ and $\phi_{upd}(\cdot)$ are two transformations in the MPNN $\phi_\theta$.
Specifically, $\phi_{msg}(\cdot)$ computes the message received by node $v_i$ from its neighbors, while $\phi_{upd}(\cdot)$ updates the node embedding by the messages.

\textbf{Transformers:}
Transformers have demonstrated remarkable success across diverse fields, achieving high performance in natural language processing~\cite{devlin-etal-2019-bert} and computer vision~\cite{dosovitskiy2021an} applications. 
Transformers employ an encoder-decoder architecture, 
where the building block is the self-attention~\cite{vaswani2017attention} mechanism.
The self-attention module operates on a sequence of $n$ tokens, $\BH \in \mathbb{R}^{n\times d}$, which can be represented by a transformation function $\phi_\theta(\cdot) : \mathbb{R}^{n \times d} \rightarrow \mathbb{R}^{n \times d}$, where $d$ is the feature dimension of each token. 
The formulation of function $\phi_\theta(\cdot)$ is shown in Equation~\eqref{eq:attention}:
\begin{align}
\BQ = \BH \BW_{Q}, \BK = \BH \BW_{K}, \BV = \BH \BW_{V}, \quad
\BA = \text{Softmax}\left(\frac{\BQ\BK^{T}}{\sqrt{d}}\right),  \quad
\Tilde{\BH} = \BA \BV, \label{eq:attention}
\end{align}
where the input $\BH$ is first transformed into query, key and value matrices, $\BQ, \BK, \BV \in \mathbb{R}^{n\times d}$, by linear weight matrices $\BW_{Q}, \BW_{K}, \BW_{V} \in \mathbb{R}^{d\times d}$, respectively. 
Then, the attention matrix $\BA \in \mathbb{R}^{n \times n}$ is computed by the inner product of the query and key, followed by the normalization of a $\text{Softmax}(\cdot)$ function.
Here, $\BA_{ij} \in [0, 1]$ indicates the influence of $\BH_j$ on $\BH_i$.
The attention matrix is finally applied to the value matrix to generate new embeddings of the tokens $\Tilde{\BH} \in \mathbb{R}^{n \times d}$. 
By interpreting the attention matrix $\BA$ in~\cref{eq:attention} as an alternative adjacency matrix, the self-attention mechanism can be viewed as a variant of MPNNs with a fully connected adjacency matrix as Equation~\eqref{eq:atten_as_mpnn:message}:

\begin{equation}
    \label{eq:atten_as_mpnn:message}
    \m_{ij} = \BA_{ij} \BV_{j}, \quad
    \BH_{i} = \sum_{j=0}^{n} \m_{ij}.
\end{equation}

After applying self-attention, Transformers employ element-wise feed-forward networks (FFN), which comprise two linear layers with ReLU activation:
\begin{equation}
    \text{FFN}(\Tilde{\BH}) = \text{max}(0, \Tilde{\BH}\BW_{1} + b_{1})\BW_{2} + b_{2},
\end{equation}
where $\BW_{1} \in \mathbb{R}^{d\times d_{1}}, \BW_{2} \in \mathbb{R}^{d_{1}\times d}$ are learnable parameters. In addition, the sublayers in the FFN will incorporate residual connections and layer normalization~\cite{ba2016layer}.

In practice, multi-head attention (MHA) is widely utilized in Transformers to enhance representation learning. 
Precisely, the $\BQ, \BK, \BV$ matrices obtained from~\cref{eq:attention} are divided into $H$ independent heads, denoted as $\BQ^{(h)}$, $\BK^{(h)}$ and $\BV^{(h)}$, respectively. 
The MHA mechanism computes representations as:
\begin{align}
    \label{eq:mha}
    \Tilde{\BH} = ||_{h=1}^{H} \text{Softmax}\left(\frac{\BQ^{(h)}\BK^{(h)T}}{\sqrt{d_h}}\right)\BV^{(h)},
\end{align}
where $d_h = d/H$ represents the dimension assigned to each head.
MHA acquires representations from various perspectives within each head and concatenates the embeddings from each head to the final representation.

Another critical component in Transformers is positional encoding, which is designed to encode the relative position of tokens in a sequential input. 
The original paper of Transformer~\cite{vaswani2017attention} introduces a parameter-free sinusoidal PE. Nevertheless, subsequent advancements, such as learned PE~\cite{gehring2017convolutional} and rotary PE~\cite{su2024roformer}, have demonstrated that the choice of PE significantly impacts the performance of Transformers, highlighting its importance in Transformers.

\section{Architectures}
\label{sec.architectures}

\begin{figure}[t]
    \centering
    \begin{minipage}{0.69\linewidth}
        \centering
        \includegraphics[width=\linewidth]{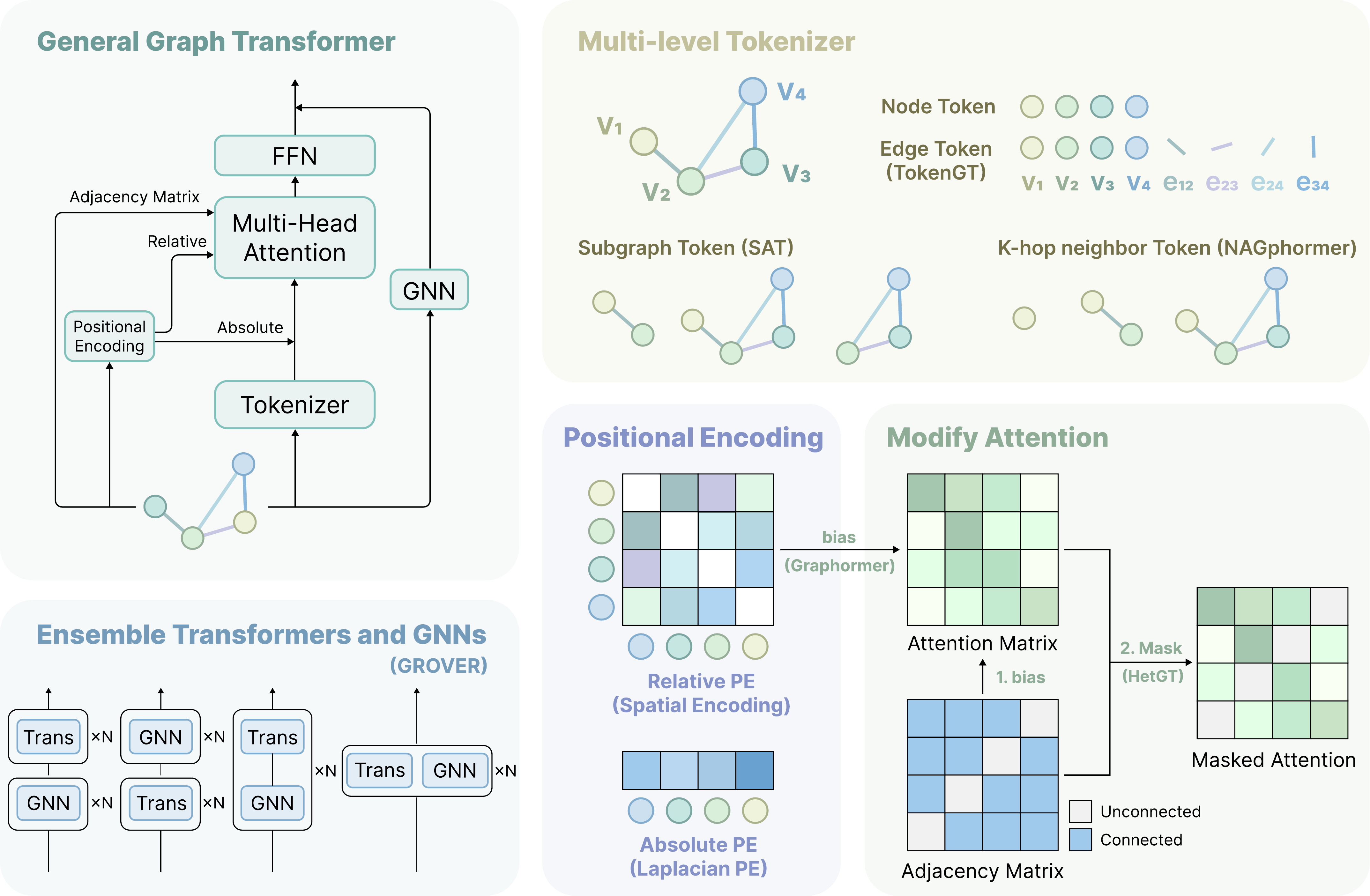}
        \caption{The overview of the architecture of Graph Transformers. The general GT part outlines the various methodologies employed to incorporate structural priors within GTs, including multi-level tokenization, positional encoding, modifying attention matrix and ensemble with GNNs. Other four parts delineate how these methodologies are applied to GTs. Methods in the parentheses are representative implementations in their corresponding taxonomies.}
        \label{fig:framework}
    \end{minipage}
    \hfill
    \begin{minipage}{0.3\linewidth}
        \centering
        \captionof{table}{A summary of existing architecture of GTs. For each model, we document the graph tokenization, the implementation of positional encoding, the utilization of structure-aware attention, and the inclusion of an additional MPNN. \textup{Abbr.,} PE: Positional Encoding, Attn: Attention, Ens: ensemble, Sub.: SubGraph, K-hop: K-hop Neighbor, RW: Random Walk, Deg.: Degree, Lap: Laplacian.}
        \vspace{-3ex}
        \resizebox{1.0\columnwidth}{!}{\begin{tabular}{l|lllll}
        \toprule
        \textbf{Model}            & \textbf{Token} & \textbf{PE} & \textbf{Attn} & \textbf{Ens}  \\
        \midrule
        SE(3)-Transformer~\cite{fuchs2020se3transformers}      &  Node   &    & \cmark      &
        \\\hline
        Graphormer~\cite{Graphormer}      &  Node   &  Deg.  & \cmark      &
        \\\hline
        Dwivedi ~\emph{et al.} ~\cite{dwivedi2021generalization}      &  Node   &  Lap. & \cmark      &
        \\\hline
        SAN~\cite{kreuzer2021rethinking}      &  Node   &  Lap. & \cmark      &
        \\\hline
        GraphiT~\cite{mialon2021graphit}       &  Sub.   &  RW & \cmark      &
        \\\hline
        TokenGT~\cite{TokenGT}  & Edge      &  Lap.  &      &
        \\\hline
        SAT~\cite{chen2022structure} & Sub.      &  Hybrid &   \cmark   &
        \\\hline
        TorchMD-Net~\cite{tholke2022torchmd} & Node      &    &   \cmark   &
        \\\hline
        GMT~\cite{min2022masked} & Node      &    &   \cmark   &
        \\\hline
        GraphGPS~\cite{rampavsek2022recipe} & Node      &  Hybrid  &      & \cmark
        \\\hline
        GKAT~\cite{choromanski2022block} & Node      &    &    \cmark  &
        \\\hline
        NodeFormer~\cite{wu2022nodeformer} & Node   &    &   \cmark   &
        \\\hline
        EGT~\cite{hussain2022global} & Edge   &  SVD  &   \cmark   &
        \\\hline
        Exphormer~\cite{shirzad2023exphormer} & Node      & Lap.   &  \cmark    &
        \\\hline
        NAGphormer~\cite{NAGphormer} & K-hop      &    &      &
        \\\hline
        GRIT~\cite{ma2023graph}      & Node    &   Deg. &       &
        \\\hline
        Graph ViT/MLP-Mixer~\cite{he2023generalization}      & Sub.    &  RW  &       &
        \\\hline
        SGFormer~\cite{wu2023sgformer}      &  Node   &    &       & \cmark
        \\\hline
        Graphormer-GD~\cite{zhang2023rethinking}      &  Node   &    &  \cmark     &
        \\\hline
        Equiformer~\cite{liao2023equiformer}      &  Node   &    &  \cmark     &
        \\\hline
        EquiformerV2~\cite{equiformer_v2}      &  Node   &    &  \cmark     &
        \\\hline
        Polynormer~\cite{deng2024polynormer}     &  Node   &  Lap.  &   \cmark    &
        \\\hline
        CoBFormer~\cite{xing2024less}      &  Sub.   &    &       & \cmark
        \\\hline
        TGT~\cite{hussain2024triplet}      &  Edge   &    &   \cmark    &
        \\\hline
        GraphGPT~\cite{GraphGPT}      &  Edge   &    &       & \cmark \\\hline
        DUALFormer~\cite{zhuodualformer}      &  Node   &    &       & \cmark\\\hline
        Primphormer~\cite{he2025primphormer}      &  Node   &    &       & \cmark \\\hline 
        ParaFormer~\cite{yuan2026paraformer}      &  Node   &    &       & \cmark \\
        \bottomrule
        \end{tabular}}
        \label{tab:architecture}
    \end{minipage}
\end{figure}

Vanilla Transformer is essentially a special GNN, where self-attention mechanisms across all nodes are operated in a fully-connected graph~\cite{muller2024attending}. 
Since Transformers ignore the original graph structure, the core objective of GT is to incorporate edge information into the Transformer architecture. 
Based on the approaches of incorporating this structural prior, in this section, we systematically categorize existing GTs into four classes:
1) Multi-level Graph Tokenization (Section~\ref{architecture:token}). 2) Structural Positional Encoding (Section~\ref{architecture:PE}). 3) Structure-aware Attention Mechanisms (Section~\ref{architecture:attention}). 4) Model Ensemble between GNNs and Transformers (Section~\ref{architecture:Ensemble}). 
Following the categorization, \cref{fig:framework} provides an overview of the architectures, while \cref{tab:architecture} summarizes representative design choices. Under the unified framework in \cref{fig:framework}, this section focuses on the architectural perspective: \cref{tab:architecture} records which structural priors are incorporated by each GT, Section~\ref{sec.theory} analyzes the expressive capability implied by these choices, and Section~\ref{sec.applications} reviews their use across different graph organization forms.
Apart from novel model architectures, practical applications of GTs necessitate high scalability or equivariance. Therefore, 
we also discuss the advanced progress in enhancing scalability (Section~\ref{architecture:Scalability}) and achieving equivariance (Section~\ref{architecture:Equivariance}) for GTs.

\subsection{Multi-level Graph Tokenization}
\label{architecture:token}

In the realm of GTs, tokenization plays a crucial role in transforming graph data into a sequential format for processing by Transformers. 
Distinguished from conventional text-based Transformers where tokenization is trivial, graph data presents unique challenges due to its structural complexity. This section explores four distinct levels of tokenization in GTs from fine-grained to coarse-grained levels, i.e., node-level, edge-level, hop-level, and subgraph-level tokenization.

\subsubsection{Node-level Tokenization}

Node-level tokenization is the most granular approach for tokenization~\cite{Graphormer,10.1145/2939672.2939754}, which
treats each node in the graph as an individual token. Furthermore, the node involved in attention can be selected through methods such as contrastive learning~\cite{chen2024leveraging}.
This approach is particularly effective when models focus on node-specific features or when the graph's topology is less critical than the attributes of individual nodes. 
By capturing detailed information about each node, node-level tokenization is well-suited for tasks such as node classification. 

\subsubsection{Edge-level Tokenization}

Edge-level tokenization extends the concept of tokenization to the connections between nodes~\cite{TokenGT,GraphGPT}. Here, each edge in the graph is treated as a token, making this approach ideal for tasks where interactions between nodes are of primary interest. Edge-level tokenization can capture the dynamics of these interactions, which is essential for tasks like link prediction or understanding the flow of information across the graph. By focusing on edges, edge-level tokenization can highlight the importance of connectivity patterns in the graph.

\subsubsection{Subgraph-level Tokenization}
Subgraph-level tokenization treats an induced local patch as the token representation~\cite{chen2022structure}, so each token summarizes the internal arrangement of nodes and edges inside that patch rather than only the attributes of one center node.
This design is effective when the topology of a local structure is itself informative, for example when motifs, cycles, clusters, or other higher-order patterns matter.
Notably, the nodes in subgraphs corresponding to different tokens are allowed to overlap, enabling a flexible and comprehensive representation of the graph.

\subsubsection{Hop-level Tokenization}
Hop-level tokenization keeps a central anchor node and constructs an ordered token sequence from its neighborhoods at different radii~\cite{NAGphormer, zhang2020graph, NAGphormer+, chen2026rethinking}. Its key signal is therefore the distance-to-anchor and the progression from 1-hop to 2-hop to higher-hop context, rather than the full identity of one induced subgraph.
Therefore, hop-level tokenization should not be simply regarded as a special case of subgraph-level tokenization with a different $k$: subgraph tokens emphasize the internal topology of a patch, while hop tokens emphasize how information is organized across multiple distance shells.
Moreover, instead of performing quadratic-complexity self-attention across all nodes, hop-level tokenization applies self-attention to each node's hop sequence. This design enables mini-batch training and makes the model scalable to large graphs.

\subsection{Structural Positional Encoding}
\label{architecture:PE}

In standard Transformers, the positional encoding module indicates the position of tokens in a sequence. To extend this module to GTs, 
 it is natural to develop methods for representing the positional embeddings of nodes in a graph. Since edge-level, hop-level, and subgraph-level tokenization approaches already incorporate structural information, in this section, we assume that the PE methods are applied at the \emph{node-level}.

Existing PE methods can be categorized into the \emph{absolute PE} and the \emph{relative PE}. 
Similar to standard Transformers, the absolute PE assigns a unique positional embedding to each node.
This embedding, either learned or parameter-free, is subsequently aggregated or concatenated with the original embedding of the node. 
In contrast, the relative PE focuses on capturing pair-wise relationships between nodes and directly applies to  the attention matrix. Therefore, we defer the discussion on relative PE to the next section. 

The main objective of the absolute PE can be formulated as utilizing a function $f$ to
extract the underlying structural information from the graph, typically from the adjacency matrix $\BA$. \cref{eq:absolute_pe} shows the usage of absolute PE as:
\begin{align}
    \label{eq:absolute_pe}
    \BP = f(\BA), \quad
    \Tilde{\BH} = g(\BH, \BP).
\end{align}
Here, the function $f$ extracts the absolute PE $\BP \in \mathbb{R}^{n\times d_p}$, where $n$ denotes the number of nodes, and $d_p$ represents the dimension of the positional embedding for each node. The function $g$ integrates absolute PE with the original node features $\BH$, either by concatenation or by employing an MLP to align the dimensions of $\BP$ and $\BH$ before summing them.

Laplacian PE leverages the eigenvectors and eigenvalues obtained from the decomposition of the Laplacian matrix as PE. The decomposition of the Laplacian can be expressed as:
\begin{equation}
    \label{eq:decomposition:laplacian}
    \BU \mathbf{\Lambda} \BU^{T}=\BI-\BD^{-1 / 2} \BA \BD^{-1 / 2},
\end{equation}
where $\BD$ denotes the degree matrix, $\BI$ is the identity matrix, $\mathbf{\Lambda}$ and $\BU$ represent eigenvalues arranged in a diagonal matrix and eigenvectors. Since the sign of pre-computed eigenvectors is arbitrary, the Laplacian PE approaches randomly adjust the sign of eigenvectors during the training stage. The first Laplacian PE~\cite{dwivedi2021generalization} proposes to utilize the $k$ smallest non-trivial eigenvectors of a node as its PE.
Another work SAN~\cite{kreuzer2021rethinking} introduces a learned Laplacian PE. For a given node $v_j$, SAN uses $\{\lambda_i$, $\BU_{ij}\}_{i=0}^{m}$ as input features for neural networks to learn the PE of node $v_j$, where $m$ is a hyperparameter  determining the number of eigenvectors considered.

Despite the effectiveness of Laplacian PE, 
it faces two underlying challenges:
1) Non-unique Eigendecompositions.
There are different  eigendecompositions of the same Laplacian. If a vector $\v$ is an eigenvector, then $-\v$ is also an eigenvector. There are non-unique solutions for eigenvectors with the multiplicities of eigenvalues. 2) Sensitivity to Perturbations. Minor perturbations in the graph structure can significantly affect the result of eigenvectors, leading to  considerable instability in Laplacian PE.

To address the first challenge in Laplacian PE, SignNet ~\cite{lim2023sign} introduces a sign-invariant network, $f$, which operates on eigenvectors as:
\begin{equation}
    f(\BU_1, \ldots, \BU_k) = \rho\left(||_{i=1}^k[\phi(\BU_i) + \phi(-\BU_i)] \right), \label{eq:signnetwork}
\end{equation}
where $\rho$ and $\phi$ are neural networks. This formulation ensures that the neural network remains invariant embeddings to the sign of the eigenvectors. 
Moreover, to tackle the occurrence of multiple eigenvector choices when there are repeated eigenvalues in the Laplacian matrix, BasisNet~\cite{lim2023sign} proposes a method to extract consistent PE from these matrices.

To address the challenge of stability in Laplacian PE, Stable and Expressive PE (SPE)~\cite{huang2024on} is introduced,
which is formulated as: 
\begin{equation}
    \BP(\BU,\mathbf{\Lambda})= \rho\big(\BU(\,\phi_1(\mathbf{\Lambda}))\BU^{\top}, \BU(\phi_2(\mathbf{\Lambda}))\BU^{\top}, ..., \BU(\phi_m(\mathbf{\Lambda}))\BU^{\top}\;\big),
    \label{eq:spe}
\end{equation}
where the input consists of the $k$ smallest eigenvalues $\lambda$ and their corresponding eigenvectors $\BV$. Rather than implementing a strict division of eigensubspaces, SPE utilizes a weighted aggregation of eigenvectors that is contingent upon the eigenvalues to ensure stability.

Singular Value Decomposition (SVD) PE~\cite{hussain2022global} provides a broader  scope of applications compared to Laplacian PE, 
as it can handle directed and weighted graphs. 
The SVD PE is computed by:
\begin{align}
  \mathbf{A} \stackrel{\mathrm{SVD}}{\approx}
  \mathbf{U}\mathbf{\Sigma}\mathbf{V}^T 
  = (\mathbf{U}\sqrt{\Sigma}) \cdot (\mathbf{V}\sqrt{\Sigma})^T
  = \hat{\mathbf{U}}\hat{\mathbf{V}}^T,
  \BP = \hat{\mathbf{U}} \parallel \hat{\mathbf{V}},
\end{align}
where $\mathbf{U,V}\in \mathbb{R}^{n\times r}$ contain the $r$ left and right singular vectors as their respective columns, each associated with the highest $r$ singular values in the diagonal matrix $\mathbf{\Sigma} \in \mathbb{R}^{r \times r}$. Similar to  Laplacian PE, SVD PE involves the random sign flipping of eigenvectors during the training phase. Consequently, building on the concept of SignNet, developing a sign-invariant SVD PE could be a potential direction for future research.

Random Walk PE (RWPE)~\cite{dwivedi2022graph} represents a PE derived from the diffusion process of a random walk. The RWPE for a node $v_i$ can be mathematically expressed in \cref{eq:rwpe} through a k-step random walk:
\begin{align}
\BP_i = \left[ \begin{array}{c} \BM_{ii}, \BM_{ii}^2, \cdots, \BM_{ii}^k \end{array} \right]\in\mathbb{R}^{k}, \label{eq:rwpe} 
\end{align}
where $\BM = \BA\BD^{-1}$ represents the random walk operator. Distinguished from Laplacian PE, RWPE does not suffer from the sign ambiguity. Under the condition that each node possesses a unique $k$-hop topological neighborhood for a sufficiently large $k$, RWPE provides a distinct node representation. As a potential future study, researchers can explore the replacement of random walk diffusion with alternative graph diffusion processes to derive PE.

Graphormer~\cite{Graphormer} introduces a heuristic method that leverages node degrees for centrality encoding. Specifically, each node is assigned a learnable vector based on its degree, which is then incorporated into the node features as the input layer, indicated as:
\begin{align}
\label{eqn:degree_encoding}
    h_i^{(0)} = x_i + z^-_{\text{deg}^{-}(v_i)} + z^+_{\text{deg}^{+}(v_i)},
\end{align}
where $z^{-}, z^{+} \in \mathbb{R}^d$ represent learnable embedding vectors defined respectively by the indegree $\text{deg}^{-}(v_i)$ and the outdegree $\text{deg}^{+}(v_i)$, respectively. For undirected graphs, $\text{deg}^{-}(v_i)$ and $\text{deg}^{+}(v_i)$ are  simplified to $\text{deg}(v_i)$. By incorporating centrality encoding into the query and key components of the attention mechanism, Graphormer enhances the ability of attention to effectively recognize both the importance and relationships among nodes.

The degree information can also be injected as post-processing. For instance, GRIT ~\cite{ma2023graph} updates the node representation after Transformer via:
   \begin{align}
   \label{eq:grit}
    \x^{\text{out}'}_i :=  \x^\text{out}_i \odot \btheta_1 + \left( \log(1 + \BD_i) \cdot \x^\text{out}_i \odot \btheta_2 \right) , 
   \end{align}
where $\btheta_1, \btheta_2$ are learnable weights.
Similarly, SAT~\cite{chen2022structure} also incorporates degree information into the residue connection as: $\x^{\text{out}'}_i=\x_{i}+1/\sqrt{\BD_i} \,\x^{\text{out}}_i$.

In the following, we elaborate on structure-aware attention mechanisms, which contain the complete implementation of relative PE.
\subsection{Structure-aware Attention Mechanisms}
\label{architecture:attention}

\begin{table}[htbp]
  \centering
  \caption{The summary of the structure-aware attention mechanism methods. $\e_{ij}$ is the edge features, $c_{ij}$ represents the shortest path between node i and node j, $\BM = \BA \BD^{-1}$ and $\BD$ is the degree matrix, $\BR$ indicates the resistance distance matrix and $\BL$ is the Laplacian matrix.}
  \vspace{-3ex}
    \resizebox{1.0\textwidth}{!}{\begin{tabular}{l|l|l|l|l|l|l}
    \toprule
          & \multicolumn{3}{c|}{\textbf{Attention Bias}} & \multicolumn{3}{c}{\textbf{Attention Mask}} \\
    \midrule
    \textbf{Method} & Graphormer~\cite{Graphormer} & Graphormer-GD~\cite{zhang2023rethinking} & GRIT~\cite{ma2023graph} & GT~\cite{dwivedi2021generalization} & HetGT~\cite{yao-etal-2020-heterogeneous} & GraphiT~\cite{mialon2021graphit} \\
    \midrule
    \textbf{Bias/Mask Term} & MLP($\e_{ij}$)+$c_{ij}$ & MLP($\BD_{ij}$) & $||_{k=n}(\BM_{ij}^{k})$ & MLP($\e_{ij}$) & if $\BA^{G}_{ij}=1$, MLP($\e_{ij}$), o.w., 0 & $e^{- \beta \BL}$ \& $(\BI  - \gamma \BL)^{p}$ \\
    \bottomrule
    \end{tabular}
    }
  \label{tab:structural-attention}
\end{table}

In Transformer blocks, the attention matrix governs the interactions between nodes, while tokenization and absolute PE augment node embeddings. 
These augmented embeddings enable Transformers to incorporate structural prior into attention mechanisms. 
In this vein, direct modification of the attention matrix is a more forthright approach for capturing  structural inductive bias. In this section, for clarity of the paper, we present all structure-aware attention mechanisms in their single-head formulation.

Adjusting the attention matrix begins with  capturing pairwise node interactions in the graph. To this end, GT leverages the graph structure to first generate a structure matrix that encodes node connectivity patterns. This structure matrix can be integrated into the attention matrix in three ways: by attention bias, by attention mask, and by edge-level tokenization. The most prevalent approaches that integrate structural information into the attention mechanism are summarized in \cref{tab:structural-attention}.
In the rest of this section, we elaborate on the three ways, respectively.
\subsubsection{Structure Matrix as Attention Bias}
By attention bias, the structural information is incorporated into the attention mechanism by adding a bias matrix $\b \in \mathbb{R}^{n\times n}$ into the inner product of the query and key matrices. 
\cref{eqn:bias} shows a general form for computing the attention matrix:
\begin{align}
\label{eqn:bias}
    \BA=\text{Softmax}\left(\frac{(\BH\BW_{Q})(\BH\BW_{K})^T}{\sqrt{d}} + \b\right),
\end{align}
where the attention bias $\b$ is specified by different approaches, which is essentially a relative PE. 
Relative PE is computed from the graph structure, aiming to understand pair-wise interactions between nodes. We define a relation matrix $\hat{\BP} \in \mathbb{R}^{R\times R}$ as the attention bias $\b$. 
$\BP_{ij}$ is determined by the function $\phi(\BH_i, \BH_j, e_{ij})$, which encodes the relationships between any pair of nodes, utilizing their embeddings $\BH_i$, $\BH_j$, and optionally incorporating edge embedding $e_{ij}$.

Graphormer~\cite{Graphormer} introduces the shortest path distance (SPD) of a shortest path $\text{SP}_{ij}=[e_1,e_2,...,e_N]$ connecting $v_i$ to $v_j$ into the attention mechanism. Graphormer incorporates two types of attention bias. The first spatial bias $\phi_(v_i, v_j)$ encodes the length of $\text{SP}_{ij}$ and the second, edge encoding $c_{ij}$, is to aggregate the edge embeddings in $\text{SP}_{ij}$. Consequently, the attention mechanism that incorporates structural information can be expressed by:
\begin{align}
\label{eqn:rnpe}
    \BA_{ij}=\text{Softmax}\left(\frac{(\BH_i\BW_{Q})(\BH_j\BW_{K})^T}{\sqrt{d}} + b_{\phi(v_i,v_j)} + c_{ij}\right), \quad
    c_{ij} =\frac{1}{N}\sum_{n=1}^{N} x_{e_n}(w^{E}_{n})^T, 
\end{align}
where $b_{\phi(v_i,v_j)}$ is a learnable scalar indexed by $\phi(v_i,v_j)$ and remains consistent across all layers.  $x_{e_n}$ and $w^{E}_{n}$ denote the feature of the $n$-th edge $e_n$ in $\text{SP}_{ij}$ and its corresponding weight, respectively. Furthermore, based on SPD, HDSE~\cite{luo2024enhancing} introduces hierarchical distance structural encoding, capturing multi-scale structural distances and  significantly enhances the Transformer's ability to model complex topologies. \textcolor{black}{While SPD is expressive, its $O(n^3)$ precomputation is often prohibitive for large graphs. HubGT~\cite{liao2026hubgt} addresses this by introducing a novel hub labeling-based indexing, which allows efficient SPD-based relative PE on million-scale graphs with significantly reduced computational overhead}

Nevertheless, Graphormer-GD~\cite{zhang2023rethinking} identifies that SPD is incapable of adequately distinguishing certain perturbations in the graph structure. To address this limitation, Graphormer-GD introduces a more robust relative PE based on resistance distance (RD). 
The attention matrix with this relative PE is represented in:
\begin{align}
\label{eq:attention-GD}
    \BA = \phi_1(\mathbf{R})\odot \operatorname{Softmax}\left(\BH \BW_Q(\BH \BW_K)^\top+\mathbf \phi_2(\mathbf{R})\right),
\end{align}
where $\mathbf{R}\in\mathbb R^{n\times n}$ represents the distance matrix with $\BR_{ij}=\{|\text{SP}_{ij}|, \text{SP}_{ij}\}$. 
Theoretical analysis demonstrates that RD-WL exhibits superior  discriminative power compared to SPD-WL, for differentiating non-isomorphic distance-regular graphs.

\textcolor{black}{Extending attention bias to domain-specific physics, Si-GT~\cite{hu2026sigt} introduces an Intra-Inter Net (IIN) attention mechanism for integrated circuit analysis. Instead of relying solely on distance metrics, it applies distinct attention biases to differentiate intra-net connections (representing wire resistance along a signal path) and inter-net connections (representing coupling capacitance between adjacent wires). This demonstrates how attention bias can effectively encode heterogeneous physical interactions.}

GRIT~\cite{ma2023graph} introduces an approach to learning relative PE by the initialization of random walk probabilities as:
$\BP_{i,j} = [\mathbf{I}, \BM, \BM^2, \dots, \BM^{K-1}]_{i,j} \in \mathbb{R}^K$,
where $\BM = \BA\BD^{-1}$ denotes the transition probability matrix of the random walk. The initialization of $\BP$, combined with MLP processing, is proven to approximate SPD. Additionally, graph-diffusion Weisfeiler-Lehman (GD-WL) with $\BP$ is strictly better than GD-WL based on SPD.

\subsubsection{Structure Matrix as Attention Mask}\label{sec:attention-mask}
An alternative approach to incorporating structure-aware attention is to perform element-wise multiplication between an attention matrix and a masking matrix, instead of treating the structure matrix as an attention bias. 
This approach can be formally expressed as: 
\begin{align}
\label{eqn:mask}
    \BA=\text{Softmax}\left(\frac{(\BH\BW_{Q})(\BH\BW_{K})^T}{\sqrt{d}} \odot \BM\right),
\end{align}
where $\BM \in \mathbb{R}^{n\times n}$ represents the masking matrix, which can be the adjacency matrix or another matrix encoding the graph structure. 

To integrate edge information into the mask matrix of a GT~\cite{dwivedi2021generalization}, $\BM_{ij}$ is defined based on the edge feature. If a connection exists, $\BM_{ij}=\BW_{E} \e_{ij}$, where $e_{ij}$ denotes the edge feature connecting node $v_i$ and node $v_j$, and $\BW_{E}$ represents a learnable weight. $\BM_{ij}$ will be set to $-\infty$ if node $v_i$ and node $v_j$ are not connected. 
This attention mask operates similarly to attention bias, as both of them only change the attention values between connected nodes.

Another classical choice for masked matrix $\BM$ is defined by the adjacency matrix 
, where $\BM_{ij}=0$ if nodes $v_i$ and $v_j$ are not connected.
By truncating attention values for disconnected nodes,  the attention is forced to focus on local neighboring nodes. Although this may reduce a GT to a GNN regarding capturing the local neighborhood information, GMT~\cite{min2022masked} and HetGT~\cite{yao-etal-2020-heterogeneous} employ distinct attention masks across different heads, thereby compelling the GT to learn from different perspectives of the graph structure. 

Similar to the attention bias, relative PE can also be used as attention masks.
GraphiT~\cite{mialon2021graphit} proposes positive definite kernels on graphs as relative PE for attention masks. Specifically, GraphiT exploits the diffusion kernel $\BM = e^{- \beta \BL}$ and the p-step random walk kernel $\BM = (\BI  - \gamma \BL)^{p}$,
where $\BL$ is the Laplacian matrix, $\beta$ and $p$ are hyperparameters. 
GraphiT demonstrates the effectiveness of these relative PEs as attention masks across various datasets.

Despite the effectiveness of classic graph diffusion, scaling it with standard attention mechanism to large graph is challenging due to the quadratic complexity regarding the number of nodes. Furthermore, as attention masks, relative PEs are not directly applicable to low-rank linear Transformers which do not explicitly construct an attention matrix. To this end, GKAT~\cite{choromanski2022block} proposes a novel Random Walks Graph-Nodes Kernel (RWGNK) with sub-quadratic complexity. RWGNK operates as  a low-rank masking directly on the query, key and value matrices in attention mechanism, 
bypassing the explicit computation of the attention matrix and thus avoiding quadratic complexity.

To unify attention bias and attention mask, EGT~\cite{hussain2022global} designs a framework which incorporates both of them as a general formula in:
\begin{align}
\label{eqn:mask_egt}
    \BA=\text{Softmax}\left(\frac{(\BH\BW_{Q})(\BH\BW_{K})^T}{\sqrt{d}} + \BE_{e}\right) \odot \BG_{e},
\end{align}
where $\BG_{e}, \BE_{e} \in \mathbb{R}^{n\times n}$ are edge embeddings generated by linear transformations.

\subsubsection{Edge-level Token as Attention Input}
Edge-level attention can be exploited in two ways. The first way focuses on computing attention only by edge tokens to generate enhanced edge representations, which are subsequently fused with node embeddings using the techniques discussed earlier.
The second way incorporates edge and node tokens simultaneously into the attention mechanism to develop structure-aware attention. 
We briefly introduce the technical essence of the representatives by the first way, i.e., EGT~\cite{hussain2022global} and TGT~\cite{hussain2024triplet}, and those by the second way, i.e., TokenGT~\cite{TokenGT} and Edgeformers~\cite{jin2023edgeformers} as below.

EGT~\cite{hussain2022global} introduces attention bias and mask from edges to nodes. In addition, the edge features can also be updated from the attention matrix in each layer as:
\begin{align}
\label{eqn:mask_egt_edge}
    \BE_{e}=f\left(\frac{(\BH\BW_{Q})(\BH\BW_{K})^T}{\sqrt{d}} + \BE_{e}\right),
\end{align}
where $f$ is a learnable function with feed-forward layers and layer normalization.

TGT~\cite{hussain2024triplet} employs triplet edge interactions to further update edge embeddings. 
Given that  $\e_{ij}$, $\e_{ik}$ and $\e_{jk}$ denote the edge embeddings of the three edges $(v_i, v_j)$, $(v_i, v_k)$ and $(v_j, v_k)$ in a triangle, the query, key and value vectors are computed by linear projections on $\e_{ij}$, $\e_{ik}$ and $\e_{jk}$ respectively. Then, 
the triplet attention computes the attention matrix and update edge embeddings:
\begin{align}
   \BA_{ijk} = \text{Softmax}_k\left(\frac{1}{\sqrt{d}}\mathbf{q}_{ij} \cdot \mathbf{k}_{jk} + b_{ik}\right) \times \sigma_1(g_{ik}) , \quad 
   \mathbf{e}_{ij} = \sigma_2\left(\sum_{k=1}^N \BA_{ijk} \mathbf{v}_{jk}\right),
   \label{eqn:tgt_at_in1}
\end{align}
where $\BA_{ijk}$ denotes the attention weight that  edge $(v_i, v_j)$ allocates to edge $(v_j, v_k)$. 
 $\sigma_1$ and $\sigma_2$ are two MLPs,  and $b_{ik}, g_{ik}$ are two scalars derived from MLP transformations on $\e_{ik}$. 

TokenGT~\cite{TokenGT} computes the attention across all the nodes and edges by concatenating the input matrix as  $\hat{\BH} = \BH || \BE$, then the calculation of the attention matrix can be represented as:
\begin{align}
\label{eqn:edge_attention}
    \BA=\text{Softmax}\left(\frac{(\hat{\BH}\BW_{Q})(\hat{\BH}\BW_{K})^T}{\sqrt{d}}\right). 
\end{align}
To additionally incorporate the structural information, in TokenGT, 
the embedding of node $v$ is combined with a node identifier $\BP_v \in \mathbb{R}^d$ by the concatenation $[\BH_v, \BP_v, \BP_v, \BT^{\mathcal{V}}]$, and the embedding of each edge $(u, v) \in \mathcal{E}$ is augmented as $[\BE_{(u,v)}, \BP_u, \BP_v, \BT^{\mathcal{E}}]$.
Here, $\BT^{\mathcal{V}}$ and $\BT^{\mathcal{E}}$ are two learnable identifiers to distinguish node and edge. 
Similar to PE, node identifiers $\BP$ can be defined via PE such as Laplacian PE.

Edgeformers~\cite{jin2023edgeformers} consist of two distinct variants: Edgeformer-E and Edgeformer-N, specializing in capturing edge and node embeddings, respectively. In this framework, edges are represented as textual data comprising multiple tokens. 
Specifically, Edgeformer-E combines these edge tokens with the tokens of their associated nodes as the input, and processes the input using self-attention.
Unlike using the entire graph as input, Edgeformer-N analyzes the ego-graph centered on node $v$. It employs Edgeformer-E to model each edge incident to $v$, and then applies an aggregation function to generate the final node representation $\BH_v$.

\subsection{Model Ensemble between GNNs and Transformers}
\label{architecture:Ensemble}
The most straightforward approach for designing GTs involves strategic combinations of GNNs and Transformers, leveraging both local structure patterns and global contextual relationships. As illustrated in \cref{fig:framework}, these ensemble architectures can be systematically divided into four categories based on the relative positioning of the GNN and Transformer blocks:
1) Sequential GNN-to-Transformer: feed the output from a GNN into a Transformer.
2) Sequential Transformer-to-GNN: feed the output from a Transformer into a GNN.
3) Interleave GNN and Transformer blocks.
4) Parallel GNN and Transformer:  feed the graph into GNN and Transformer concurrently, and fuse the output representations into one representation.

In the first category, GTs initially process the graph by inputting it into a GNN, which can be regarded as tokenizing the graph at the subgraph level. The GNN aggregates information from local neighborhoods to refine node embeddings. Then, the augmented node embeddings are fed into a Transformer, enabling the model to learn from subgraph tokens, as discussed in Section~\ref{architecture:token}.

The second category of architectures is commonly employed when Transformer blocks have been pretrained. For instance, in the domain of protein data, Transformers~\cite{lin2023evolutionary} have demonstrated effective capabilities in capturing amino acid residue  representations. 
Protein GT frameworks typically exploit pretrained Transformers to generate an initial node representation, followed by the refinement through a GNN regarding the spatial graph structure.
A more comprehensive discussion will be provided in Section~\ref{sec:application-geometric}.

Interleaving GNN and Transformer blocks, as the third category of model ensemble, is a simple yet effective architecture.
For example, Mesh Graphormer~\cite{lin2021mesh} interleaves GNN and Transformer blocks to reconstruct human poses, and GROVER~\cite{rong2020self} adopts a hybrid strategy of combining GNN and Transformer to learn molecular representations.

By paralleling GNN and Transformer, GTs can adaptively learn the importance of both local and global information.
GraphGPS~\cite{rampavsek2022recipe} utilizes the parallel architecture which combines the outputs from a MPNN and a Transformer. In addition, GraphGPS leverages MPNN to update the edge embeddings, which can be utilized to further update the PE.

SGFormer~\cite{wu2024simplifying} theoretically proves that a single-layer attention is sufficiently expressive to capture the global interactions among nodes. Accordingly, SGFormer proposes a simplified GT architecture that incorporates a single-layer self-loop linear attention mechanism alongside GCN blocks. By combining the final representations from the Transformer and GNN, SGFormer exhibits considerable scalability and competitive performance in node property prediction tasks.

CoBFormer~\cite{xing2024less} aims to address the issue of over-globalization in GTs. To this end, CoBFormer parallelizes GCN blocks with the Transformer and proposes a collaborative training strategy to supplement the local graph structure knowledge from GCN into the Transformer. Specifically, CoBFormer incorporates an additional loss function to align the output representations of the GCN and Transformer, thereby enabling mutual supervision between the two modules.

\subsection{Towards Scalability in Graph Transformer}
\label{architecture:Scalability}
Recall that the self-attention mechanism in Transformers introduces a quadratic computational complexity regarding the number of nodes. Since real-world graphs may contain millions or even billions of nodes,  Transformers often struggle to scale to large graphs efficiently.
Thus, designing efficient attention mechanisms for large-scale graphs remains a significant challenge for the scalability of GTs. 

To reduce the complexity of the attention mechanism to linear, one most straightforward approach is to integrate  GNNs with linear Transformers. For example,  GraphGPS~\cite{rampavsek2022recipe} adopts established Transformers that utilize linear attention mechanisms, e.g.,  combining Performer~\cite{choromanski2021rethinking} and BigBird~\cite{zaheer2020big} with other GNN modules. Nonetheless, experiments on GraphGPS reveal that although linear attention mechanisms improve scalability, they tend to degrade performance.
SGFormer~\cite{wu2024simplifying}, an alternative ensemble-based GT,
introduces a linear attention mechanism with self-loop propagation.
Theoretical analysis demonstrates that a single-layer attention is sufficient to capture global interactions, enabling SGFormer to achieve scalability and competitive accuracy in node classification tasks.
Another linear GT, Polynormer~\cite{deng2024polynormer} implements a local-to-global attention mechanism, which learns high-degree polynomial from input features, including node features and graph structure.

CobFormer~\cite{xing2024less} presents a bi-level global attention module aimed at mitigating the over-globalization issue while simultaneously reducing model complexity. Initially, CobFormer partitions the entire graph into clusters. Subsequently, a bi-level attention mechanism operates at both the intra-cluster and inter-cluster levels, which reduces memory consumption significantly. 
Similarly, Polynormer~\cite{deng2024polynormer} introduces a linear framework by polynomial network, where each output element is represented as a polynomial function of the input features. To enable permutation equivariance and combine local and global information, it calculates local attention on neighboring nodes and global attention on the entire graph as the coefficients in the polynomial network.

A notable limitation of the structural attention mechanism, as discussed in Section \ref{architecture:attention}, is its difficult applicability to linear attention. This stems from the fact that linear attention mechanisms do not explicitly construct an attention matrix, making it challenging to incorporate structural information through attention bias or attention mask. To this end, NodeFormer~\cite{wu2022nodeformer} introduces an edge-level regularization loss as Equation~\eqref{eqn-loss-mle} that encourages attention values between connected nodes in a graph are close to 1.0, represented as:
\begin{equation}\label{eqn-loss-mle}
    \mathcal L_{e}(\BA,  \BA^{G}) = - \frac{1}{NL} \sum_{l=1}^L \sum_{(u, v)\in \mathcal E} \frac{1}{d_u} \log \BA_{uv}^{(l)},
\end{equation}
where $L$ denotes the total number of layers in NodeFormer, and $d_u$ represents the degree of node $u$. 
Since this loss function only requires computations over edges, NodeFormer efficiently manages the complexity of edge regularization at $\mathcal{O}(|\mathcal E|)$, maintaining the overall model complexity as $\mathcal{O}(|\mathcal V| + |\mathcal E|)$.

Exphormer~\cite{shirzad2023exphormer} incorporates a sparse attention mechanism to achieve linear complexity. In essence, this sparse attention mechanism combines the adjacency matrix, expander graph, and virtual node. An expander graph randomly connects nodes and ensure that each node maintains an equal degree, resulting in the expander possessing a number of edges that is linear in relation to the nodes. Despite its linear complexity, the expander graph preserves the spectral approximation of a complete graph. Moreover, Exphormer achieves competitive performance compared to dense attention. SP\_Exphormer~\cite{shirzad2024even} further pushes this direction by exploring more aggressive sparsification of graph attention. Similarly, ANS-GT~\cite{zhang2022hierarchical} samples subgraphs with different strategies and employs graph coarsening to reduce the computational complexity.

An alternative approach avoids feeding the entire graph into the Transformer. NAGphormer~\cite{NAGphormer} transforms the $k$-hop neighborhood $\mathcal{N}^{k}(v)$ into a neighborhood embedding $\mathbf{x}^k_v$ using an aggregation operator $\phi$. This aggregated embedding is then treated as a token within the Transformer, enabling the model to learn the embedding of node $v$.
By aggregating neighboring nodes before processing by Transformer, NAGphormer circumvents the need to input a large number of nodes into the Transformer. Moreover, NAGphormer+~\cite{NAGphormer+} enhances the feature of $\mathbf{x}^k_v$ by randomly masking a portion of neighbors to achieve better performance. 
In addition, VCR-Graphormer~\cite{fu2024vcrgraphormer} then employs random walk to rewire the graph by virtual nodes. By maintaining a graph with virtual nodes, VCR-Graphormer controls its complexity, keeping it linear with respect to the number of nodes while still capturing long-range dependencies.

\subsection{Geometric Graph Transformers}
\label{architecture:Equivariance}
Given the wide range of real-world scientific applications involving GTs, the study of geometric GTs is crucial for modeling 3D graph data,  such as molecular systems and protein structures.
The core design principle of these frameworks lies in ensuring the 3D invariance and/or equivariance of the model. 
This section briefly reviews the state-of-the-art  equivariant GTs that have been successfully applied to 3D graph modeling.

The most straightforward approach to learn the structural relationships is incorporating the 3D relative distance as an additional edge embedding, which remains unchanged under Euclidean transformations.
For example, Graphormer~\cite{Graphormer} introduces spatial encoding, where an MLP is used to encode the relative distance between atoms, effectively capturing structural relationships. 
This paradigm has demonstrated its efficacy in various frameworks for learning molecular representations~\cite{zhou2023unimol}. 
Additionally, other invariant features, such as the angle between edges~\cite{yan2022periodic}, can be included to represent orientation information. These invariant features are usually encoded using kernel functions, such as the Radial Basis Function~\cite{choudhary2021atomistic}, to enhance the model's expressivity.

TorchMD-Net~\cite{tholke2022torchmd} represents another equivariant model that incorporates the interatomic distance $r_{ij}$ into its framework. The process begins by projecting $r_{ij}$ into two distinct multidimensional filters, denoted as $\BD^{K}$ and $\BD^{V}$, using the following expressions:
\begin{equation}\label{torchmd}
    \BD^{K} = \sigma_1(r_{ij}), \quad \BD^{V} = \sigma_2(r_{ij}),
\end{equation}
where $\sigma_1$ and $\sigma_2$ are two MLPs. Subsequently, 
TorchMD-Net replaces the traditional Softmax function with the SiLU function to compute the attention matrix as shown in Equation~\eqref{eq:torchmd:attention}:
\begin{equation}
    \label{eq:torchmd:attention}
    \BA = \text{SiLU}((\BH\BW_{Q})(\BH\BW_{K})^T \BD^{K}) \cdot \phi(\d_{ij}),
\end{equation}
where $\phi$ denotes a cutoff function that assigns the value of $0$ whenever $\d_{ij}$ exceeds a predefined threshold. The final representation is then computed by:
\begin{equation}
    \BZ = \sigma_3(\BA \BV \BD^{V}),
\end{equation}
where $\sigma_3$ represents another learnable linear transformation.

For tasks such as conformation generation, where the model needs to generate atomic coordinates, Uni-Mol~\cite{zhou2023unimol} proposes a simple SE(3)-equivariant head, represented as:
\begin{align}
    \vec{\BX}_i = \vec{\BX}_i + \frac{1}{n} \sum_{j=1}^{n} \left(\vec{\BX}_i - \vec{\BX}_j\right)c_{ij}.
\end{align}
where $c_{ij}$ represents the learned relationship embedding between node $v_i$ and $v_j$. 
To improve efficiency, Uni-Mol updates the coordinates only in the final layer of the model.

GVP-Transformer~\cite{hsu2022learning} represents an encoder-decoder framework based on the Transformer architecture, designed for the task of protein inverse folding. The model is structured to intake protein structures and subsequently generate corresponding protein sequences. As an encoder, GVP-Transformer utilizes the GVP-GNN~\cite{jing2021learning}, capable of extracting features that are translation-invariant and rotation-equivariant, to effectively model protein structures. This is followed by the application of a Transformer decoder to produce valid protein sequences.

\textcolor{black}{Beyond single-conformer modeling, FACET~\cite{nguyen2026facet} introduces a scalable framework for conformer ensemble learning. It leverages a differentiable GT to approximate the Fused Gromov-Wasserstein (FGW) distance, enabling efficient and geometry-aware aggregation of multiple 3D conformations.}

Examples of high-order steerable GTs include SE(3)-Transformer~\cite{fuchs2020se3transformers}, Equiformer~\cite{liao2023equiformer}, Equi-formerV2~\cite{equiformer_v2}, \textcolor{black}{TetraGT~\cite{feng2026tetragt}, Q-GT~\cite{quan2026pseudoriemannian} and EquiformerV3~\cite{liao2026equiformerv3}}. These models employ equivariant attention mechanisms utilizing higher-degree representations of steerable features~\cite{han2024survey}, which fall beyond the focus of this survey.

\section{Theories}
\label{sec.theory}
Beyond the empirical effectiveness of GTs, it is also important to understand their theoretical foundations.
This section first reviews the expressive capability of GTs from the perspectives of tokenization and positional encoding (Section~\ref{theory:expressiveity}), and then discusses the relationships between GTs and other graph learning paradigms (Section~\ref{theory:relationship}).

\subsection{Expressivity}
\label{theory:expressiveity}
Following the order of Section~\ref{sec.architectures}, we discuss the expressivity of structural tokenization and positional encoding. These two components play different roles: tokenization determines \emph{what structural objects are treated as tokens for attention}, whereas PE provides \emph{additional structural relations among these tokens}. Therefore, the theoretical issue is not only whether a GT is stronger, but also what kinds of graph ambiguities it can distinguish and at what computational or statistical cost.

\subsubsection{Structural Tokenization}
Tokenization determines the space on which a Transformer operates. Node- and edge-level tokenization remain close to primitive graph entities, and their expressivity therefore still depends on additional structural bias introduced by PE, edge features, or architectural constraints. By contrast, higher-order tokenization directly exposes multi-node objects to attention. TokenGT~\cite{TokenGT} is a representative example showing that the token inventory itself can improve the theoretical expressivity of a pure Transformer beyond a vanilla node-token view.

This can be further explained by the framework of~\cite{expressive-token}. If a Transformer is given suitable \(k\)-tuple input tokens \(\mathbf{X}^{(0,k)} \in \mathbb{R}^{n^k \times d}\), then its \(t\)-th layer can emulate the \(t\)-th iteration of a \(k\)-order WL procedure. This result indicates that stronger-than-1-WL behavior does not arise from attention alone, but from allowing the model to attend over higher-order objects or relations.

\textcolor{black}{It is also useful to distinguish \emph{subgraph-level} and \emph{hop-level} tokenization, although they are often discussed together. Subgraph-level tokenization treats an induced patch as one object~\cite{chen2022structure}. Its identity depends on the internal arrangement of nodes and edges inside the patch, and is therefore suitable when motif type, cycle structure, or other higher-order local configurations are important. Hop-level tokenization instead keeps a central anchor node and organizes its context into an ordered sequence of neighborhoods indexed by distance~\cite{NAGphormer, zhang2020graph, NAGphormer+}. Its key inductive bias is radial decomposition: the model learns how evidence is distributed across the 1-hop, 2-hop, \ldots shells around an anchor, rather than learning the identity of one patch as a whole. In this sense, hop-level tokenization should not be simply regarded as subgraph-level tokenization with a different \(k\); the former emphasizes multi-hop propagation patterns and scalability, whereas the latter emphasizes the isomorphism type of a local structure.}

This distinction also helps explain when \(\ge 2\)-WL signals are necessary. Such signals are especially useful when the target depends on relations among multiple nodes that 1-WL-style node updates cannot distinguish, such as symmetric roles, motif participation, or local structures sharing the same multiset of neighbor features. In these cases, subgraph-, edge-, or tuple-aware tokenization can expose the missing structure more directly. By contrast, when node or edge attributes already break these symmetries, or when the task mainly depends on short-range local evidence, higher-order tokenization may introduce additional computation without bringing clear gains. This also explains why many node-level GTs, and even strong MPNN baselines, remain competitive on practical benchmarks despite weaker worst-case expressivity.

\subsubsection{Positional Encoding} \label{sec.pe-theory}
PE does not change the token itself, but provides structural relations for the attention layers. From this perspective, absolute PE assigns each token a structural coordinate, typically derived from spectra, diffusion, or degree statistics. Relative PE keeps the signal in pairwise form and injects it directly into attention, allowing the model to compare two tokens through shortest-path distance, resistance distance, random-walk affinity, or other relation scores.

This distinction leads to several important theoretical observations. Black et al.~\cite{black2024comparing} show, through a conversion based on 2-equivariant graph networks~\cite{maron2018invariant}, that on graphs without node features the graph-distinguishing power of absolute and relative PE can be equivalent. Once node features are present, however, converting relative PE into absolute PE can lose information. Intuitively, compressing a pairwise relation into two separate node-wise labels is weaker than exposing that relation directly to the attention mechanism. Therefore, relative PE can be more suitable when pairwise structure plays a central role.

This comparison also indicates what PE can and cannot replace. Stronger absolute PE helps break graph symmetries and provides node-level tokens with a more informative global coordinate system. Stronger relative PE determines which pairwise structural tests can be performed directly in attention. However, even a powerful PE usually enriches a model with unary or pairwise signals, rather than arbitrary subgraph identities. In other words, PE can compensate for node-level tokenization only when the task is mainly governed by node positions or pairwise relations; if the key evidence lies in motif identity or higher-order local topology, higher-order tokens remain the more direct choice.

\textcolor{black}{Fundamentally, PE and Tokenization serve orthogonal roles in GTs. Tokenization defines the \emph{observation space} (the entities the model can attend to), while PE establishes the \emph{coordinate system} within that space. If the downstream task fundamentally requires reasoning over high-order topological structures (e.g., specific chemical motifs), imposing a powerful spectral coordinate system (PE) onto a node-level observation space remains sub-optimal. In such cases, explicitly shifting the observation space via subgraph-level or edge-level tokenization is a mathematically more direct and empirically more efficient approach.}

Recent theoretical studies also help explain the gap between expressive power and empirical performance~\cite{li2024what}. First, WL-style analyses concern worst-case distinguishability, whereas benchmark accuracy is also influenced by optimization, sample efficiency, and supervision alignment. Second, the theoretically strongest positional signals are not always the most robust in practice: Laplacian-based PE suffers from sign and basis ambiguity, spectral features can be sensitive to perturbations, and some relative PEs are difficult to combine with sparse or linear attention. Third, many datasets already contain rich node attributes or mainly local targets, so the cases separating 1-WL from stronger tests may not dominate the final loss. 

Therefore, a theoretically stronger PE does not uniformly translate to empirical superiority~\cite{grotschla2026benchmarking}; researchers must navigate a strict trade-off between expressive power, computational budget, and perturbation stability, as summarized in Table~\ref{tab:pe_tradeoffs}. Practical gains appear only when the task truly requires the additional structural distinctions and the model can exploit them within its computation and sample budget.

\begin{table*}[htbp]
\centering
\caption{Theoretical and Practical Trade-offs of Graph Positional Encodings.}
\vspace{-3ex}
\label{tab:pe_tradeoffs}
\renewcommand{\arraystretch}{0.7}
\resizebox{\textwidth}{!}{
\begin{tabular}{l c c c l}
\toprule
\textbf{PE Strategy} & \textbf{Expressivity Limit} & \textbf{Complexity} & \textbf{Stability} & \textbf{Applicability Constraints} \\
\midrule
Degree & 1-WL & $\mathcal{O}(|E|)$ & High & Fails on regular graphs \\
Laplacian PE & $>$ 1-WL (Spectral) & $\mathcal{O}(|V|^3)$ & Low & Sign/Basis ambiguity; Undirected graphs \\
SVD PE & $>$ 1-WL (Spectral) & $\mathcal{O}(|V|^2 r)$ & Moderate & Supports directed/weighted graphs \\
Random Walk PE & Subgraph-aware & $\mathcal{O}(k |E|)$ & High & Depends on unique $k$-hop neighborhoods \\
Shortest Path (Rel.) & $>$ 1-WL (Distance) & $\mathcal{O}(|V|^3)$ & High & Hard to adapt to linear attention \\
\bottomrule
\end{tabular}
}
\vspace{-2ex}
\end{table*}

{
\subsubsection{Beyond Pairwise Graphs: Signed, Directed, and Hypergraph Extensions}
\label{theory:extensions}

The four graph organization forms surveyed in Section~\ref{sec.applications} focus primarily on undirected, pairwise graphs with optional attributes. Several natural extensions, such as directed graphs, signed graphs, and hypergraphs, are worth noting, as they can be addressed within the same theoretical framework without requiring a fundamentally new taxonomy.

\textbf{Directed graphs} are already covered by the relational graph form. Knowledge graphs and AMR graphs, which appear in Section~\ref{sec:application-relational}, are inherently directed: edges encode orientation-specific semantics (e.g., \texttt{subject}$\to$\texttt{predicate}$\to$\texttt{object}). The same attention-bias and edge-type-specific projection mechanisms discussed for relational graphs directly handle directionality.

\textbf{Signed graphs}, where edges carry positive or negative signs, also fit the relational framework without modification. Social trust/distrust networks and user-rating graphs are the typical instances. From the perspective of a GT, a signed edge is simply a typed relation with values in $\{+1,-1\}$; the same attention-bias machinery and edge-type encoding covered in Section~\ref{sec.architectures} applies. The key design choice is whether the sign modulates the attention score (as in signed GNNs) or is treated as a separate token-level attribute.

\textbf{Hypergraphs} incorporates hyperedge connecting an arbitrary set of nodes rather than a pair. Example applications include co-authorship networks, session-based recommendation, and protein complex modeling. From the GT expressivity perspective, a hyperedge is precisely the $k$-tuple input studied in TokenGT~\cite{TokenGT} and the $k$-WL emulation framework of~\cite{expressive-token}: encoding a $k$-ary relation as a higher-order token lifts the model beyond 1-WL expressivity. This is conceptually consistent with the higher-order tokenization row of Table~\ref{tab:guidance-matrix}, where hypergraph-structured data would directly benefit from subgraph-, edge-, or tuple-aware tokenization. 

In summary, all three extensions map naturally onto the architectural categories and theoretical frameworks already surveyed in Sections~\ref{sec.architectures} and~\ref{theory:expressiveity}, reinforcing the generality of the organization-form and component-frequency perspective.
}

\subsection{Relationship with Other Graph Learning Methods}
\label{theory:relationship}
The characteristics of GTs can be elucidated through being compared with other graph learning methods. In this section, we examine studies that compare GT with MPNN, graph structure learning, and graph attention networks (GATs).

\subsubsection{MPNN}
Compared with MPNNs, GTs integrate self-attention mechanisms and PE. A recent study~\cite{li2024what} shows that self-attention can improve the convergence rate of GTs, while PE helps identify the core neighborhood for each node and thus improves generalization. GTs with shortest-path distance as relative PE can also be theoretically more expressive than classical MPNNs~\cite{black2024comparing}. However, stronger theoretical expressivity does not necessarily imply a uniform empirical advantage.

An alternative approach to infuse global information into each node is to introduce a virtual node connected to all nodes in a graph. 
Despite the simplicity of this idea, MPNN with the virtual node~\cite{cai2023connection} surprisingly serves as a strong baseline in Long Range Graph Benchmark~\cite{dwivedi2022long}. A recent study~\cite{rosenbluth2024distinguished} further shows that no single algorithm consistently outperforms the others when comparing GTs and MPNNs with a virtual node.

In addition, the over-smoothing problem~\cite{rong2019dropedge}, characterized in deep MPNNs, also exists in Transformers~\cite{shi2022revisiting}, leading to indistinguishable node embeddings in deeper layers.
As Transformer is a special form of Graph Attention Networks (GAT)~\cite{velivckovic2017graph}, it shares the same over-smoothing phenomenon as GAT, 
leading to an exponential degeneration of expressive power regarding the number of layers. 
To mitigate over-smoothing, ParaFormer~\cite{yuan2026paraformer} proposes a Generalized PageRank Attention mechanism to preserve the diverse frequency information in the graph structure from the perspective of graph signal processing.

\subsubsection{Graph Structural Learning}
Graph Structure Learning (GSL) is closely related to GTs, which aims to automatically 
refining graph structures when the input graph is noisy or incomplete, or inferring implicit graph structures when explicit graph structure is unavailable~\cite{GSLB}, in a parameterized way.
GTs can be regarded as a special form of GSL, achieved by self-attention that learns a fully connected `soft' graph structure~\cite{mp-all-the-way-up}. 
By utilizing attention-oriented techniques, such as the attention mask in \cref{sec:attention-mask} and discrete structure sampling in NodeFormer~\cite{wu2022nodeformer}, the learned graph structure can be sparsified to reflect real-world topology.
\section{Applications}
\label{sec.applications}

\tikzstyle{leaf}=[mybox,minimum height=1em,
fill=hidden-orange!40, text width=20em, text=black,align=center,font=\tiny,
inner xsep=2pt,
inner ysep=1pt,
]

\colorlet{linecolor}{purple!60}

GTs have been adopted across many domains. In this section we review the landscape through the lens of graph organization form relational, geometric, dynamic, and heterogeneous or multimodal. This grouping cuts down on repetitive descriptions of similar task setups and ties downstream use cases more directly to the architectural components covered in Section~\ref{sec.architectures}. Figure~\ref{fig:app-taxonomy} gives an overview of the representative tasks and models under each form.

\subsection{A Unified View of Graph Organization Forms}
\label{sec:application-unified}

Under the four forms, a large fraction of GT applications can be captured by the shared task templates:
\begin{align}
    y &= \phi_\theta(\mathcal G, \mathcal X), \\
    s &= \phi_\theta(\mathcal G_a, \mathcal X_a, \mathcal G_b, \mathcal X_b), \\
    \hat{\mathcal G}, \hat{\mathcal X} &= \phi_\theta(\mathcal G, \mathcal X, c), \\
    \hat{\BT} &= f(\phi_\theta(\mathcal G, \mathcal X)).
\end{align}
Here, $\mathcal G$ denotes the graph structure, $\mathcal X$ denotes optional side information such as geometry, time, or modality-specific features, $y$ is a prediction target, $s$ is a score for a graph pair, $(\hat{\mathcal G}, \hat{\mathcal X})$ denotes generated or refined graph content, and $\hat{\BT}$ is decoded text. Based on this formulation, we are able to summarize the dominant structural signals and the corresponding GT design choices across different graph organization forms.

\begin{figure*}[t]
\centering
\resizebox{1.0\textwidth}{!}{
\begin{forest}
  for tree={
  forked edges,
  grow=east,
  draw=linecolor,
  line width=0.9pt,
  reversed=true,
  anchor=base west,
  parent anchor=east,
  child anchor=west,
  base=middle,
  font=\normalsize,
  rectangle,
  rounded corners,
  align=left,
  minimum width=2.5em,
  inner xsep=4pt,
  inner ysep=1pt,
  },
  where level=1{text width=7.5em}{},
  where level=2{text width=9em,font=\normalsize}{},
  where level=3{font=\normalsize, fill=pink!20}{},
  [GT Applications, rotate=90,edge=linecolor
        [Relational (\textsection~\ref{sec:application-relational}),text width=8em,edge=linecolor
            [Prediction \& Generation, text width=10em,edge=linecolor
                [MAT~\cite{maziarka2020molecule}{,}
                R-MAT~\cite{maziarka2022relative}{,}
                GROVER~\cite{rong2020self}{,}
                LiGhT~\cite{li2022kpgt}{,}
                CoAtGIN~\cite{zhang2022coatgin}{,}
                BatmanNet~\cite{wang2024batmannet}{,}
                DMP~\cite{zhu2023dual}{,}
                MolSpectra~\cite{MolSpectra}{,}\\
                DiGress~\cite{vignac2023digress}{,}
                CDGS~\cite{huang2023conditional}{,}
                Uni-Mol~\cite{zhou2023unimol}{,}
                Uni-Mol+~\cite{lu2024data}{,}
                Uni-Mol2~\cite{ji2024exploring}{,}
                Mudiff~\cite{hua2024mudiff}{,}
                JODO~\cite{huang2024learning}{,}
                GTMGC~\cite{xu2024gtmgc}{,}
                GraphDiT~\cite{liu2024graph}{,}\\
                BrainNetTF~\cite{kan2022brain}{,}
                Cai et al.~\cite{cai2022graph}{,}
                THC~\cite{dai2023transformer}{,}
                TSEN~\cite{hu2023transformer}{,}
                ALTER~\cite{yu2024longrange}{,}
                BioBGT~\cite{peng2025biologically}
                ,text width=54em,edge=linecolor
                ]
            ]
            [Graph-to-Text \& \\ Reasoning, text width=10em,edge=linecolor
                [Zhu et al.~\cite{zhu-etal-2019-modeling}{,}
                Cai et al.~\cite{cai-lam-2020-graph}{,}
                HetGT~\cite{yao-etal-2020-heterogeneous}{,}
                ASAG~\cite{agarwal2022multi}{,}
                GraphFormer~\cite{yang2021graphformers}{,}
                KGTransformer~\cite{kgtransformer}{,}
                Relphormer~\cite{Relphormer}{,} \\
                TG-Transformer~\cite{zhang2020text}{,}
                KG-R3~\cite{retrieval-read}{,}
                GT-BEHRT~\cite{poulain2024graph}
                ,text width=54em,edge=linecolor
                ]
            ]
        ]
        [Geometric \& \\ Periodic (\textsection~\ref{sec:application-geometric}),text width=8em,edge=linecolor
            [Property Prediction, text width=10em,edge=linecolor
                [Equiformer~\cite{liao2023equiformer}{,}
                EquiformerV2~\cite{equiformer_v2}{,}
                TorchMD-NET~\cite{tholke2022torchmd}{,}
                SE(3)-Transformer~\cite{fuchs2020se3transformers}{,}
                GNS-TAT~\cite{zaidi2023pretraining}{,}
                TGT~\cite{hussain2024triplet}{,}
                TransFun~\cite{boadu2023combining}{,} \\
                HEAL~\cite{gu2023hierarchical}{,}
                scMoFormer~\cite{tang2023single}{,}
                Stability Oracle~\cite{diaz2024stability}{,}
                ProstT5~\cite{10.1093/nargab/lqae150}{,}
                Saprot~\cite{su2024saprot}{,}
                Matformer~\cite{yan2022periodic}{,}
                CrystalFormer~\cite{wang2024conformal}{,}\\
                CrysGraphFormer~\cite{sun2024crysgraphformer}{,}
                DPA-2~\cite{zhang2023dpa}{,}
                MatterSim~\cite{yang2024mattersim}{,}
                OMat24~\cite{barroso2024open}{,}
                ComFormer~\cite{yan2024complete}{,}
                Mesh Graphormer~\cite{lin2021mesh}{,}
                PoseGTAC~\cite{zhu2021posegtac}{,} \\
                GTRS~\cite{zheng2022lightweight}{,}
                Graformer~\cite{zhao2022graformer}{,}
                3DMOTFormer~\cite{ding20233dmotformer}{,}
                SGraFormer~\cite{zhang2024deep}{,}
                SGFormer~\cite{lv2024sgformer}
                ,text width=54em,edge=linecolor
                ]
            ]
            [Interaction \& Docking, text width=10em,edge=linecolor
                [GraphSite~\cite{yuan2022alphafold2}{,}
                RTMScore~\cite{shen2022boosting}{,}
                IGT~\cite{liu2022improved}{,}
                GeoT~\cite{morehead2022geometric}{,}
                GGT~\cite{chen2023gated}{,}
                GraphormerDTI~\cite{gao2024graphormerdti}{,}
                AttentionMGT-DTA~\cite{wu2024attentionmgt}{,}\\
                GTAMP-DTA~\cite{tian2024gtamp}{,}
                Graph-BERT~\cite{jha2023graph}{,}
                HGIN~\cite{zhao2023geometric}{,}
                Uni-Mol~\cite{zhou2023unimol}{,}
                GeoDock~\cite{chu2024flexible}{,}
                EBMDock~\cite{wu2024ebmdock}
                ,text width=54em,edge=linecolor
                ]
            ]
            [Generation \& Design, text width=10em,edge=linecolor
                [JODO~\cite{huang2024learning}{,}
                MUDiff~\cite{hua2024mudiff}{,}
                GTMGC~\cite{xu2024gtmgc}{,}
                Uni-Mol~\cite{zhou2023unimol}{,}
                Uni-Mol+~\cite{lu2024data}{,}
                Uni-Mol2~\cite{ji2024exploring}{,}
                GVP-Transformer~\cite{hsu2022learning}{,}
                LM-Design~\cite{zheng2023structure}{,} \\ 
                ProRefiner~\cite{zhou2023prorefiner}{,}
                FAIR~\cite{zhang2023fullatom}{,}
                PocketGen~\cite{zhang2024efficient}
                ,text width=54em,edge=linecolor
                ]
            ]
        ]
        [Dynamic (\textsection~\ref{sec:application-dynamic}),text width=8em,edge=linecolor
            [Forecasting, text width=10em,edge=linecolor
                [Social Attention~\cite{vemula2018social}{,}
                Trajectron~\cite{ivanovic2019trajectron}{,}
                TrafficPredict~\cite{ma2019trafficpredict}{,}
                STAR~\cite{yu2020spatio}{,}
                SSAGCN~\cite{lv2023ssagcn}{,}
                Trafformer~\cite{jin2023trafformer}{,}
                LLGformer~\cite{jin2025llgformer}{,}
                IGT~\cite{zhou2023inductive}{,}\\
                GMAN~\cite{zheng2020gman}{,}
                HS-GT~\cite{fan2021heterogeneous}{,}
                HST-GT~\cite{zhao2023hst}{,}
                GCT-TTE~\cite{mashurov2024gct}{,}
                HPST-GT~\cite{wang2025hpst}
                ,text width=54em,edge=linecolor
                ]
            ]
            [Propagation \& Events, text width=10em,edge=linecolor
                [StA-HiTPLAN~\cite{khoo2020interpretable}{,}
                DGTR~\cite{wei2023dgtr}{,}
                Lgt~\cite{xia2024lgt}{,}
                HeteroSGT~\cite{zhang2024heterogeneous}{,}
                GCNs-MT~\cite{chang2024novel}{,}
                PHAROS~\cite{nguyen2024portable}{,}
                PSGT~\cite{zhu2024propagation}{,}
                GT-BEHRT~\cite{poulain2024graph}
                ,text width=54em,edge=linecolor
                ]
            ]
        ]
        [Heterogeneous / \\ Multimodal (\textsection~\ref{sec:application-heterogeneous}),text width=8em,edge=linecolor
            [Interaction Prediction, text width=10em,edge=linecolor
                [PMGT~\cite{liu2021pre}{,}
                GMT~\cite{min2022masked}{,}
                GFormer~\cite{li2023graph}{,}
                LightGT~\cite{wei2023lightgt}{,}
                MGFormer~\cite{chen2024masked}{,}
                TransGNN~\cite{zhang2024transgnn}{,}
                SIGFormer~\cite{chen2024sigformer}{,}
                Rankformer~\cite{chen2025rankformer}
                ,text width=54em,edge=linecolor
                ]
            ]
            [Representation Learning, text width=10em,edge=linecolor
                [GNN-nested Transformer~\cite{yang2021graphformers}{,}
                Edgeformer~\cite{jin2023edgeformers}{,}
                Heterformer~\cite{jin2023heterformer}{,}
                GTP~\cite{zheng2022graph}{,}
                AMIGO~\cite{nakhli2023sparse}{,}
                MulGT~\cite{zhao2023mulgt}{,}
                MG-Trans~\cite{shi2023mg}{,} \\
                CGT~\cite{wang2024breast}{,}
                IGT~\cite{shi2024integrative}{,}
                SpaFormer~\cite{wen2023single}{,}
                HEAT~\cite{chan2023histopathology}
                ,text width=54em,edge=linecolor
                ]
            ]
            [Grounding \& QA, text width=10em,edge=linecolor
                [M-DGT~\cite{chen2022multi}{,}
                DUET~\cite{chen2022think}{,}
                Multimodal GT~\cite{he2023multimodal}{,}
                mDT~\cite{hebert2024multi}{,}
                KGEMT~\cite{zheng2024knowledge}
                ,text width=54em,edge=linecolor
                ]
            ]
        ]
    ]
\end{forest}
}
\vspace{-5ex}
\caption{Overview of GT applications under four graph organization forms.}
\label{fig:app-taxonomy}
\vspace{-4ex}
\end{figure*}

{\color{black}
\subsection{Relational Graphs with Discrete Semantics}
\label{sec:application-relational}

Relational graphs consist primarily of discrete entities and typed relations. The core difficulty in this setting is modeling non-local semantic dependencies without relying on Euclidean geometry. GTs tend to perform well here when equipped with relation-aware attention biases, shortest-path or centrality encodings, and subgraph-level context aggregation. Common instantiations include 2D molecular graphs, brain connectivity networks, AMR semantic graphs, and knowledge graphs, spanning tasks from single-graph prediction and molecular generation to graph-to-text decoding and link-based reasoning.

\subsubsection{Data Organization and Shared Task Templates}

The standard input is a graph $\mathcal G = (\mathcal V, \mathcal E)$ with node and edge attributes. The common objectives are graph-level prediction $y = \phi_\theta(\mathcal G)$, discrete graph generation $\hat{\mathcal G} = \phi_\theta(\mathcal G, c)$, and graph-conditioned decoding $\hat{\BT} = f(\phi_\theta(\mathcal G))$. Compared with geometry-aware applications, a central issue is how to expose edge types, multi-hop dependencies, and symbolic composition to the attention layers.

\subsubsection{Single-Graph Prediction and Generation}

Two-dimensional molecular modeling is a common motivation for this form. When molecules are treated as chemical graphs without explicit coordinates, GTs mainly serve to propagate information across distant functional groups and chemically meaningful paths. MAT~\cite{maziarka2020molecule} and R-MAT~\cite{maziarka2022relative} inject structural information as attention biases, while GROVER~\cite{rong2020self} and CoAtGIN~\cite{zhang2022coatgin} combine local message passing with global attention to cover both motif-level and graph-level dependencies. This relational template also covers brain connectivity analysis, where graphs are defined by correlations rather than geometry. BrainNetTF~\cite{kan2022brain}, Cai et al.~\cite{cai2022graph}, THC~\cite{dai2023transformer}, ALTER~\cite{yu2024longrange}, BioBGT~\cite{peng2025biologically} and \textcolor{black}{LGC-SGT~\cite{zhang2026localglobal}} all use GTs to learn graph-level subject representations from weighted connectivity patterns.

Discrete generation and self-supervised learning fit the same organization. For molecule generation, DiGress~\cite{vignac2023digress}, CDGS~\cite{huang2023conditional}, and GraphDiT~\cite{liu2024graph} generate or refine discrete molecular graphs while preserving graph sparsity and conditional control. GraphXForm~\cite{pirnay2025graphxform} demonstrates that GT backbones extend to computer-aided molecular design beyond standard property prediction. For pre-training, MolT5~\cite{edwards-etal-2022-translation}, GROVER~\cite{rong2020self}, LiGhT~\cite{li2022kpgt}, BatmanNet~\cite{wang2024batmannet}, and DMP~\cite{zhu2023dual} optimize masking or cross-view consistency objectives on symbolic molecular views. Transformer-M~\cite{luo2023one}, MoleBLEND~\cite{yu2024multimodal}, MolSpectra~\cite{MolSpectra} and RELGT~\cite{dwivedi2026relational} align the relational scaffold with complementary modalities during pre-training. In this setting, GTs are most effective when graph semantics depend on long relation chains or higher-order motifs; when graphs are very small and dependencies are strongly local, simpler MPNNs remain competitive.

\subsubsection{Semantic Graph-to-Text and Reasoning}

Semantic graphs such as AMR graphs and knowledge graphs instantiate the same organization form, but their outputs are text or symbolic answers rather than graph labels. For graph-to-text generation, the shared template is an encoder-decoder pipeline $\hat{\BT} = f(\phi_\theta(\mathcal G))$. Existing AMR-to-text models encode structural relations through shortest-path information and graph-specific attention biases, as in Zhu et al.~\cite{zhu-etal-2019-modeling}, Cai et al.~\cite{cai-lam-2020-graph}, and HetGT~\cite{yao-etal-2020-heterogeneous}. ASAG~\cite{agarwal2022multi} uses the same idea for student answer scoring by turning both responses into AMR graphs before graph-aware comparison.

Knowledge-graph reasoning fits this form, since the primary structural signal is still carried by typed relations. GraphFormer~\cite{yang2021graphformers} encodes shortest-path relations as attention biases, KGTransformer~\cite{kgtransformer} combines graph-aware pre-training with MoE routing for complex reasoning, and Relphormer~\cite{Relphormer} strengthens masked modeling with high-order relational signals. KG-R3~\cite{retrieval-read} further follows this template by retrieving a candidate subgraph and then applying a GT over nodes and edges to answer subject-relation queries. A common relation-centric lens thus connects superficially different tasks such as AMR generation and KG completion.

\subsection{Geometric and Periodic Graphs}
\label{sec:application-geometric}

Geometric graphs augment topology with coordinates, distances, directions, and angles; periodic graphs add lattice structure on top of that. GTs operating on these graphs must capture long-range dependencies while respecting geometric consistency, which makes pair representations, distance-aware attention, and equivariant layers far more central than in purely relational settings. Common instantiations include 3D molecular conformations, protein structures, periodic crystal materials, and geometric meshes or point clouds from vision, spanning tasks from geometric property prediction and pairwise docking to 3D structure generation and inverse design.

\subsubsection{Data Organization and Shared Task Templates}

We denote a geometric graph as $\vec{\mathcal G} = (\mathcal V, \mathcal E, \vec{\BX})$, where $\vec{\BX}$ stores coordinates or other geometric descriptors. For periodic materials, an additional lattice matrix $\BL$ is required. The dominant tasks are property prediction $y = \phi_\theta(\vec{\mathcal G})$, pair modeling $s = \phi_\theta(\vec{\mathcal G}_a, \vec{\mathcal G}_b)$, and structure generation or refinement $(\hat{\mathcal G}, \hat{\vec{\BX}}) = \phi_\theta(\vec{\mathcal G}, c)$. Since these tasks depend on geometry itself rather than only on connectivity, an important design issue is how geometry is incorporated into attention, pair features, and equivariant updates.

\subsubsection{Property Prediction on Structured Object Graphs}

3-dimensional molecular property prediction is the representative of this form. Equiformer~\cite{liao2023equiformer}, EquiformerV2~\cite{equiformer_v2}, TorchMD-NET~\cite{tholke2022torchmd}, and SE(3)-Transformer~\cite{fuchs2020se3transformers} use equivariant GT layers so that molecular predictions remain stable under rigid transformations. The same organization form appears in protein property prediction, where TransFun~\cite{boadu2023combining}, HEAL~\cite{gu2023hierarchical}, Stability Oracle~\cite{diaz2024stability}, scMoFormer~\cite{tang2023single}, Saprot~\cite{su2024saprot}, and ProstT5~\cite{10.1093/nargab/lqae150} fuse sequence-derived residue information with spatial protein graphs. GTs fit well here because residues that are distant along the backbone may still be tightly coupled in 3D space.

Crystal property prediction follows the same geometric template but adds periodicity. Matformer~\cite{yan2022periodic}, CrystalFormer~\cite{wang2024conformal}, and CrysGraphFormer~\cite{sun2024crysgraphformer} encode distance and angular information directly into attention. ComFormer~\cite{yan2024complete} combines invariant and equivariant layers, while MatterSim~\cite{yang2024mattersim} and OMat24~\cite{barroso2024open} demonstrate that large-scale pre-training on materials can further improve transferability. Geometric computer-vision tasks can also be interpreted in the same way once images are converted into meshes or point clouds. Mesh Graphormer~\cite{lin2021mesh}, PoseGTAC~\cite{zhu2021posegtac}, Graformer~\cite{zhao2022graformer}, GTRS~\cite{zheng2022lightweight}, 3DMOTFormer~\cite{ding20233dmotformer}, SGraFormer~\cite{zhang2024deep}, and SGFormer~\cite{lv2024sgformer} all operate on explicit spatial relations rather than on domain-specific semantics.

\subsubsection{Pairwise Interaction and Docking}

When the target depends on the interaction between two geometric graphs, the shared task is pair modeling. Protein-ligand affinity, protein-protein docking, and drug-target interactions all fall here, and the model must learn geometry-aware pair representations, not two independent graph embeddings. RTMScore~\cite{shen2022boosting}, IGT~\cite{liu2022improved}, GeoT~\cite{morehead2022geometric}, GGT~\cite{chen2023gated}, GraphormerDTI~\cite{gao2024graphormerdti}, AttentionMGT-DTA~\cite{wu2024attentionmgt}, and GTAMP-DTA~\cite{tian2024gtamp} all enhance attention with pairwise spatial features or pretrained biochemical embeddings. HGIN~\cite{zhao2023geometric}, Graph-BERT~\cite{jha2023graph}, and GraphSite~\cite{yuan2022alphafold2} similarly inject pairwise residue or distance information to model biological interactions. For docking, Uni-Mol~\cite{zhou2023unimol}, GeoDock~\cite{chu2024flexible}, and EBMDock~\cite{wu2024ebmdock} refine pair representations into final poses or energy-consistent structures.

\subsubsection{Structure Generation and Inverse Design}

Generation and inverse design are also important applications under this organization form. In molecular modeling, JODO~\cite{huang2024learning}, MUDiff~\cite{hua2024mudiff}, Uni-Mol~\cite{zhou2023unimol}, and GTMGC~\cite{xu2024gtmgc} generate or refine 3D conformations by coupling discrete chemical graphs with continuous coordinates. Protein design can be viewed under the same formulation, where the graph structure is given and the model predicts the amino-acid sequence or the missing pocket structure. GVP-Transformer~\cite{hsu2022learning}, LM-Design~\cite{zheng2023structure}, ProRefiner~\cite{zhou2023prorefiner}, FAIR~\cite{zhang2023fullatom}, and PocketGen~\cite{zhang2024efficient} all exploit GTs to align structural context with sequence design.

Self-supervised learning on geometric graphs also follows this organization form. Uni-Mol~\cite{zhou2023unimol}, GNS-TAT~\cite{zaidi2023pretraining}, Frad~\cite{feng2023fractional}, and SliDe~\cite{ni2024sliced} denoise coordinates, bond geometry, or force-related perturbations before downstream prediction or generation. These objectives are especially useful when downstream supervision is scarce. That said, GTs do not always beat simpler geometric GNNs: if the graph is small, interactions are strongly local, and memory is tight, a carefully tuned equivariant MPNN can be the better trade-off.

\subsection{Dynamic Graphs and Event Streams}
\label{sec:application-dynamic}

Dynamic graphs are indexed by time. The challenge here is jointly modeling graph structure and temporal evolution, and deciding how spatial and temporal signals should interact. GTs are well-suited because self-attention naturally captures long temporal dependencies, while graph-aware modules preserve local interaction structure. Common instantiations include multi-agent trajectory forecasting, traffic flow and event prediction, rumor propagation detection, and event-stream reasoning in electronic health records, spanning tasks from future-state forecasting and classification to sequence-level understanding of propagation cascades.

\subsubsection{Data Organization and Shared Task Templates}

Let $\{\mathcal G_t\}_{t=1}^{T}$ denote a sequence of graphs or event-induced graph states. The dominant tasks are forecasting $\hat{\BY}_{T+1:T'} = \phi_\theta(\{\mathcal G_t\}_{t=1}^{T})$ and sequence-level prediction $y = \phi_\theta(\{\mathcal G_t\}_{t=1}^{T})$. Compared with static graphs, important design questions include whether spatial and temporal attention should be coupled or decoupled, how event order should be encoded, and how complexity can be controlled when the observation horizon grows.

\subsubsection{Temporal Forecasting}

Traffic systems motivate this form most visibly. In trajectory prediction, the graph evolves as agents move and their interactions change over time. Social Attention~\cite{vemula2018social}, Trajectron~\cite{ivanovic2019trajectron}, TrafficPredict~\cite{ma2019trafficpredict}, and SSAGCN~\cite{lv2023ssagcn} explicitly couple spatial and temporal edges. STAR~\cite{yu2020spatio} separates spatial and temporal Transformer blocks, while Trafformer~\cite{jin2023trafformer} and LLGformer~\cite{jin2025llgformer} inject structural and temporal positional information directly into attention.

Traffic event prediction uses the same organization form but typically at a larger scale and often with heterogeneous node types. GMAN~\cite{zheng2020gman} parallelizes spatial and temporal attention, IGT~\cite{zhou2023inductive} constructs type-specific bipartite subgraphs before Transformer fusion, GCT-TTE~\cite{mashurov2024gct} combines GNN and Transformer encoders, and HS-GT~\cite{fan2021heterogeneous}, HST-GT~\cite{zhao2023hst}, and HPST-GT~\cite{wang2025hpst} extend heterogeneous GTs to temporal forecasting. These studies suggest that GTs are particularly helpful when long-range temporal dependency matters or when the model needs to fuse multiple interaction scales.

\subsubsection{Propagation and Event Understanding}

Rumor detection is another instance of dynamic graph reasoning because the prediction depends on the temporal unfolding of a propagation tree rather than on a static social graph. StA-HiTPLAN~\cite{khoo2020interpretable} models tweet relations as attention biases, DGTR~\cite{wei2023dgtr} combines structural and temporal Transformers, and Lgt~\cite{xia2024lgt} integrates GNN and Transformer components for joint local-global reasoning. HeteroSGT~\cite{zhang2024heterogeneous}, GCNs-MT~\cite{chang2024novel}, PHAROS~\cite{nguyen2024portable}, and PSGT~\cite{zhu2024propagation} further show that propagation tasks often require heterogeneous or topology-masked attention.

Event-stream reasoning also appears outside social media. GT-BEHRT~\cite{poulain2024graph}, for example, models electronic health records as patient-specific event graphs with relation-aware attention. This captures what the organization-form taxonomy buys: rumor cascades and patient histories come from different domains, but both involve sequence-aware problems on graphs and benefit from similar GT patterns. In practice, dynamic GTs need careful complexity control; when the horizon is short or interactions are nearly local, recurrent or message-passing baselines still hold up well.

\subsection{Heterogeneous and Multimodal Graphs}
\label{sec:application-heterogeneous}

The last form covers graphs that integrate multiple node types, edge types, or modalities. GTs are useful here not only for their global receptive field, but also because they can flexibly fuse tokens from structurally different objects. Type-specific projections, heterogeneous attention, cross-attention, and graph pooling are recurring design patterns. Common instantiations include user-item recommendation graphs, text-rich networks, image-induced graphs from pathology slides and cellular images, and multimodal graphs for visual grounding and question answering, spanning tasks from interaction prediction and representation learning to cross-modal inference.

\subsubsection{Data Organization and Shared Task Templates}

We write the input as $\mathcal G = (\mathcal V, \mathcal E, \mathcal M)$, where $\mathcal M$ denotes type-specific or modality-specific features. The most common objectives are interaction prediction $\hat{y}_{u,v} = \phi_\theta(\mathcal G, u, v)$, graph representation learning $z = \phi_\theta(\mathcal G)$, and cross-modal inference $\hat{y} = \phi_\theta(\mathcal G^{(1)}, \mathcal G^{(2)}, \mathcal M)$. Unlike the previous forms, the main challenge here is not only structural range, but also semantic mismatch across node types and modalities.

\subsubsection{Interaction Prediction on Bipartite and Heterogeneous Graphs}

Recommendation systems provide the canonical example. User-item interactions form bipartite or heterogeneous graphs, often augmented with textual, acoustic, or visual side information. PMGT~\cite{liu2021pre}, GMT~\cite{min2022masked}, and MGFormer~\cite{chen2024masked} use graph sampling, masking, and positional bias to learn scalable user-item representations. GFormer~\cite{li2023graph}, LightGT~\cite{wei2023lightgt}, SIGFormer~\cite{chen2024sigformer}, and TransGNN~\cite{zhang2024transgnn} combine local graph encoders with Transformer blocks to fuse structural and multimodal signals. Rankformer~\cite{chen2025rankformer} further aligns the architecture with ranking objectives, while HIRE~\cite{allinonefang} shows that attention-based heterogeneous modeling is especially useful in cold-start scenarios.

\subsubsection{Representation Learning on Text- and Image-Induced Graphs}

In many applications, a graph is first constructed from another modality and then fed to a GT for representation learning. Text-rich graphs are a canonical case: GNN-nested Transformer~\cite{yang2021graphformers} alternates local and global processing, Edgeformer~\cite{jin2023edgeformers} explicitly models textual edge content, and Heterformer~\cite{jin2023heterformer} assigns type-specific projections to text-rich versus text-free nodes. Once graph tokens carry long natural-language descriptions, heterogeneous fusion tends to matter more than pure graph topology.

Whole-slide and cellular images follow a similar pattern, with graphs induced from image patches, cells, or biomarkers rather than given a priori. GTP~\cite{zheng2022graph} uses GCNs and pooling to reduce the number of image patches before Transformer modeling; AMIGO~\cite{nakhli2023sparse}, MulGT~\cite{zhao2023mulgt}, MG-Trans~\cite{shi2023mg}, and IGT~\cite{shi2024integrative} learn on biomarker-specific or task-specific graph views; SpaFormer~\cite{wen2023single}, CGT~\cite{wang2024breast}, and HEAT~\cite{chan2023histopathology} further inject random-walk, Laplacian, spatial, and heterogeneous edge information. Compared with standard vision transformers, these models exploit the fact that the relevant context is structured by cell adjacency and tissue organization rather than by raster order.

\subsubsection{Cross-Modal Grounding, Navigation, and Question Answering}

Cross-modal tasks combine graph structure with an additional modality at inference time. M-DGT~\cite{chen2022multi} performs visual grounding by constructing a graph over image regions and conditioning it on text. DUET~\cite{chen2022think} uses graph-aware navigation memories together with textual cross-attention for embodied navigation. Multimodal GT~\cite{he2023multimodal}, mDT~\cite{hebert2024multi}, and KGEMT~\cite{zheng2024knowledge} all use semantic or structural graphs to improve multimodal question answering and social-media understanding. These tasks differ from recommendation or text-graph learning because the graph is only one part of the reasoning context; nevertheless, the same heterogeneous GT mechanisms are reused.

\subsection{Practical Guidance: What to Use When}
\label{sec:application-guidance}

Across Sections~\ref{sec:application-relational}--\ref{sec:application-heterogeneous}, a consistent pattern emerges: the most effective GT design choices are predicted by the organization form of the input graph rather than by the application domain. By tracking how often each architectural component appears in successful GT models of each form, the design options can be organized into three tiers: high frequency (nearly all strong models under this form adopt the component), medium frequency (many works include it, and it usually brings nontrivial gains), and low frequency or emerging (used by a few or very recent works and worth exploring, but not yet standard). Table~\ref{tab:guidance-matrix} summarizes this mapping.
}

\newcommand{\HI}{$\bullet$$\bullet$$\bullet$}
\newcommand{\MI}{$\bullet$$\bullet$$\circ$}
\newcommand{\LI}{$\bullet$$\circ$$\circ$}
\newcommand{\NA}{$\circ$$\circ$$\circ$}

\begin{table*}[t]
\centering
\caption{Frequency of GT architectural components across four graph organization forms, based on Sections~\ref{sec:application-relational}--\ref{sec:application-heterogeneous}. \HI\, \MI\ and \LI\ indicate high, medium and low frequency, respectively, and \NA\ means the module is not applicable in this application.}
\vspace{-3ex}
\label{tab:guidance-matrix}
\small
\renewcommand{\arraystretch}{0.7}
\begin{tabularx}{\textwidth}{>{\raggedright\arraybackslash}p{3.8cm} XXXX}
\toprule
& \textbf{Relational} & \textbf{Geometric} & \textbf{Dynamic} & \textbf{Heterogeneous} \\
\midrule
\multicolumn{5}{c}{\textit{Multi-level Graph Tokenization (Sec.~3.1)}} \\
\midrule
Node-level Tokenization   & \HI & \HI & \HI & \HI \\
Edge-level Tokenization   & \MI & \LI & \LI & \MI \\
Subgraph-level Tokenization& \MI & \LI & \LI & \LI \\
Hop-level Tokenization    & \MI & \NA & \LI & \LI \\
\midrule
\multicolumn{5}{c}{\textit{Structural Positional Encoding (Sec.~3.2)}} \\
\midrule
Shortest Path (SPD)     & \HI & \MI & \LI & \LI \\
Random Walk PE           & \HI & \LI & \LI & \LI \\
Degree PE                & \MI & \LI & \LI & \LI \\
Laplacian PE             & \LI & \MI & \LI & \NA \\
Temporal PE              & \NA & \NA & \HI & \NA \\
\midrule
\multicolumn{5}{c}{\textit{Structure-aware Attention (Sec. 3.3)}} \\
\midrule
Attn Bias / Mask         & \HI & \MI & \MI & \MI \\
Distance-based Attn      & \NA & \HI & \LI & \NA \\
Cross-Attention          & \LI & \MI & \LI & \HI \\
\midrule
\multicolumn{5}{c}{\textit{GNN-Transformer Ensemble (Sec. 3.4)}} \\
\midrule
Local GNN + Global Trans.& \MI & \LI & \MI & \HI \\
Type-specific Projections& \MI & \LI & \LI & \HI \\
\midrule
\multicolumn{5}{c}{\textit{Enhancing Scalability (Sec. 3.5)}} \\
\midrule
Sparse / Linear Attn     & \LI & \LI & \MI & \LI \\
Sampling / Pooling       & \MI & \LI & \LI & \MI \\
\midrule
\multicolumn{5}{c}{\textit{Achieving Equivariance (Sec. 3.6)}} \\
\midrule
Equivariant Updates      & \NA & \HI & \NA & \NA \\
Pairwise Geometric Feature& \NA & \HI & \LI & \NA \\
Angle Encoding           & \NA & \MI & \NA & \NA \\
\bottomrule
\end{tabularx}
\vspace{-5ex}
\end{table*}

{\color{black}
In summary, Table~\ref{tab:guidance-matrix} shows the pattern. Node-level tokenization is the universal default, while subgraph- and edge-level tokenization are most common for relational graphs. For relational graphs, the components marked \HI\ are attention bias with SPD and RWPE. For geometric graphs, \HI\ falls on distance-based attention, pairwise geometric features, and equivariant updates. For dynamic graphs, \HI\ is concentrated on temporal PE and attention bias. For heterogeneous graphs, \HI\ centers on type-specific projections, cross-attention, and local-global hybrid architectures. Components marked \LI\ indicate emerging directions that could become standard as the field matures. The table is meant as a reference to be tested, not a fixed prescription.
}

\section{Discussion}
\label{sec.future}
The rapid advancement of GTs has opened several promising avenues for future research across diverse scientific domains. In this section, we first reflect on the limitations of message-passing GNNs, then outline foundational and emerging directions for GT development.

\subsection{Reflections on the Current Message Passing Paradigm} \label{sec:reflections}

\noindent\textbf{Graph structure is treated as ground truth, not as a prior.}
The message-passing paradigm restricts information flow to the edges of the input graph, which implicitly assumes that the given structure is ideal, since  aggregation happens if and only if two nodes are connected. This assumption is reasonable for molecular and crystal graphs, but less so for constructed graphs, such as brain networks~\cite{kan2022brain} and image patch graphs~\cite{zheng2022graph}, whose structures may be noisy.  A noisy structure introduces an incorrect inductive bias into the model and degrades performance. To reduce the reliance on the structure, GTs relax this assumption through fully-connected attention, and reintroduce the structural information as a soft prior through positional encodings~\cite{dwivedi2021generalization, kreuzer2021rethinking} (Section~\ref{architecture:PE}) and attention
biases~\cite{Graphormer, zhang2023rethinking} (Section~\ref{architecture:attention}).
 
\noindent\textbf{Feature fusion may cost more than it adds.}
Neighborhood aggregation injects structural information but also mixes away the original node features. This dilution is an intrinsic property of the aggregation operator, and it underlies several of the problems commonly attributed to message passing. Over-smoothing~\cite{chen2020measuring, rong2019dropedge} is the most prominent example: what it eventually destroys is the original node features rather than the structure. Whether it is harmful depends on how much task-relevant information the original node features carry. In citation and social networks, fixed high-dimensional attributes often provide the dominant signal, making dilution harmful. By contrast, when node features are categorical, such as the atom type in molecular graphs, they index a learnable embedding table as a token does in NLP. The embedding is optimized jointly with the model, and most task-relevant information resides in the topology, so the same dilution costs far less. From this view, many studies on deep GNNs are not about depth, since DropEdge~\cite{rong2019dropedge}, the initial residuals of APPNP~\cite{appnp} and GCNII~\cite{gcnii}, and jumping knowledge networks~\cite{xu2018representation} adopt different mechanisms but all preserve the original node features. Over-smoothing and over-globalization~\cite{xing2024less} are two forms of the same failure, a node's informative local signal being washed out once by excessive neighborhood aggregation and once by excessive global attention. 
GTs alleviate the dilution through residual connections, yet how much original feature information a model ultimately retains is rarely quantified or reported by current benchmarks.

\noindent\textbf{Expressivity is measured by isomorphism, not generalization.}
The WL hierarchy is a combinatorial construction for graph isomorphism testing~\cite{expressive-token}, which examines whether two non-isomorphic graphs can be distinguished in the worst case. GIN~\cite{xu2018powerful} and Morris et al.~\cite{morris2019} first adopted this construction to measure the expressive power of GNNs, showing that message-passing GNNs are at most as powerful as the $1$-WL test, and it has remained the dominant measure since. However, this measure differs from the goal of graph learning, which is to generalize over a data distribution. A parameterized GNN is also not exactly equivalent to a WL test, for two reasons: the equivalence between GIN and 1-WL assumes a countable feature space that continuous node attributes violate, and WL results concern the existence of parameters realizing a function rather than whether optimization can find them~\cite{li2024what}. As discussed in Section~\ref{sec.theory}, a theoretically stronger model is not necessarily more accurate, because the components that raise WL power often introduce instability. GTs are analyzed by the same measure, which usefully diagnoses the structural signals to which an architecture is blind (Section~\ref{sec.pe-theory}) but does not predict empirical performance.

\subsection{Future Directions}

\textbf{Scaling Graph Transformers.}
Despite the remarkable success achieved in scaling Transformers, a question remains whether scaling up the GTs would similarly enhance performance. Uni-Mol2~\cite{ji2024exploring} scales the GT to billions of parameters, showcasing improvements on molecular downstream tasks. This scalability in Uni-Mol2 is feasible due to abundant molecular graph data. However, scaling GTs in domains with limited graph-structured data remains a significant obstacle. Innovative approaches like LM-design~\cite{zheng2023structure} utilize GNNs as structural adapters for pretrained protein language models, integrating both limited structural information and abundant sequence data from existing protein datasets. Despite these advancements, the field still lacks a comprehensive framework that effectively addresses the fundamental challenges of scaling GTs in data-constrained environments.

\textbf{Graph Transformers for Multimodal Data Modeling.} 
Integrating a GNN with a pretrained Transformer from the language domain allows the model to generate captions for graphs. MolCA~\cite{liu2023molca}, an example of cross-modal GT, employs a molecular GNN to encode molecular representations and feeds these embeddings into a Transformer decoder pretrained on language data to generate captions for the molecule. 
In the context of complex graph systems, e.g., proteins~\cite{yuan2024functional}, utilizing GTs to integrate graph structures offers an opportunity to enhance our comprehension of protein properties through pretrained language models, thereby indicating a promising direction. 

\textbf{Alternative Approaches to Capture Long-range Dependencies.} Recently, the Transformer architecture has encountered increased competition, evidenced by models like Mamba~\cite{gu2023mamba}. Notably, Graph Mamba~\cite{behrouz2024graph} has demonstrated very competitive performance when compared to GTs. As the Transformer model presents limitations, including scalability and over-smoothing issues as outlined in the survey, the success of Graph Mamba suggests the possibility of another model capable of effectively capturing global interactions within graph data. A promising direction involves the development of a model that might deviate from the self-attention paradigm and more efficiently capture global interactions, circumventing the issues inherent in  Transformer.

\textbf{Bridging Theory and Architecture Beyond Spectral Encoding.}
The existing theory of GTs has mostly focused on spectral positional encodings and the WL-expressivity of multi-level tokenization (Section~4). Other architectural components lack a corresponding theoretical treatment. Attention bias and mask mechanisms (Section~3.3) encode structure through several distinct strategies, but their representational power compared to one another has not been formally characterized. GNN-Transformer arrangements (Section~3.4), whether serial, parallel, or interleaving, are currently chosen by heuristics, without theory that predicts which arrangement benefits which type of inductive bias. Equivariant GTs (Section~3.6) rest on solid group-theoretic foundations for geometric graphs, but extending these foundations to other symmetry groups or to non-geometric forms remains unexplored.

\textbf{Graph Transformers as Graph Foundational Models.} Transformer has been a cornerstone in building foundational models across various domains, such as natural language processing and computer vision. 
However, the critical importance of graph-structured data representation in scientific modeling and social network analysis has led to significant interest in Graph Foundational Models~\cite{mao2024position}. Pioneering works such as GROVER~\cite{rong2020self} and DPA-2~\cite{zhang2023dpa}, are pretrained on molecular and crystal domains, establish a new paradigm for scientific machine learning. Current studies suggest that scalable node- or edge-aware tokenization, relative structural encodings, and sparse or hybrid attention modules are promising choices for graph foundation models
. These developments highlight the potential of GTs as fundamental building blocks for constructing next-generation Graph Foundational Models across diverse application domains.


\section{Conclusion}
\label{sec.conclusion}
In this survey, we present a comprehensive review of the recent advancements in Graph Transformers. We begin by examining strategies for incorporating structural information into the Transformer architecture, including multi-level graph tokenization, structural positional encodings, structure-aware attention mechanisms, and hybrid models that integrate GNNs with Transformers. We also discuss two prominent challenges in GT research: scalability, which pertains to improving architectural efficiency for handling large-scale graphs, and equivariance, which focuses on designing models that respect symmetry constraints inherent in specific data domains. Additionally, we explore theoretical developments to uncover connections and evaluate the expressiveness of GTs. \textcolor{black}{For applications, we reorganize the literature around four graph organization forms (relational, geometric, dynamic, heterogeneous) unified under shared task templates, showing that the input graph form predicts effective GT design choices better than the application domain alone. The Practical Guidance table (Table~\ref{tab:guidance-matrix}) maps architectural components to graph forms by adoption frequency.} 
Concluding with an exploration of prospective research avenues, this survey aims to furnish valuable insights and guidance for future investigations in the field of GTs.

\section*{Acknowledgment}
This work was jointly supported by the following projects: Damo Academy (Hupan Laboratory) through Damo Academy (Hupan Laboratory) Innovative Research Program, Research Grants Council of the Hong Kong Special Administrative Region, China (No. CUHK 14206625), Shenzhen Science and Technology Innovation Commission (JCYJ20220530143002005), Shenzhen Ubiquitous Data Enabling Key Lab (ZDSYS20220527171406015), and Tsinghua University SIGS Start-up fund (QD2022024C).

\bibliographystyle{ACM-Reference-Format}
\bibliography{sample-base}

\appendix

\end{document}